\pdfoutput=1

\documentclass[11pt]{article}

\usepackage{acl}
\usepackage{times}
\usepackage{latexsym}
\usepackage{times}
\usepackage{latexsym}
\usepackage[inkscapelatex=false]{svg}
\usepackage{multicol}
\usepackage{graphicx} 
\usepackage{natbib}  
\usepackage{caption} 
\usepackage{amssymb}
\usepackage[ruled,linesnumbered]{algorithm2e}
\usepackage{color}
\usepackage{amsmath}
\usepackage{amsthm}
\usepackage{color}
\usepackage{adjustbox}
\usepackage{multirow}
\usepackage{booktabs}
\usepackage{amsmath}
\usepackage{hyperref}
\usepackage[inkscapelatex=false]{svg}
\usepackage{subfig}
\newtheorem{definition}{Definition}
\newtheorem{theorem}{Theorem}

\newtheorem{lemma}{Lemma}
\newtheorem{Proof}{Proof}

\usepackage[T1]{fontenc}

\usepackage[utf8]{inputenc}

\usepackage{microtype}

\title{Sen2Pro: A Probabilistic Perspective to Sentence Embedding \\ from Pre-trained Language Model}
\author{Lingfeng Shen, Haiyun Jiang, Lemao Liu, Shuming Shi \\
        Natural Language Processing Center \\ Tencent AI Lab }

\begin{document}
\maketitle
\begin{abstract}
Sentence embedding is one of the most fundamental tasks in Natural Language Processing and plays an important role in various tasks.
The recent breakthrough in sentence embedding is achieved by pre-trained language models (PLMs). 
Despite its success, an embedded vector (Sen2Vec) representing a point estimate does not naturally
express uncertainty in a task-agnostic way.
This paper thereby proposes an efficient framework on probabilistic sentence embedding (Sen2Pro) from PLMs, and it represents a sentence as a probability density distribution in an embedding space to reflect both model uncertainty and data uncertainty (i.e., many-to-one nature) in the sentence representation. 
The proposed framework performs in a plug-and-play way without retraining PLMs anymore, and it is easy to implement and generally applied on top of any PLM.  
The superiority of Sen2Pro over Sen2Vec has been theoretically verified and practically illustrated on different NLP tasks. 
\end{abstract}

\section{Introduction}
\label{submission}
Sentence embedding, which maps an input sentence to a point (i.e., a vector) in an embedded space, is one of the most fundamental tasks in Natural Language Processing (NLP), and it plays an important role in various downstream tasks such as sentiment analysis, text classification, and natural language inference \cite{howard2018universal,reimers2019sentence,gao2021simcse}. There is a surge of interest in learning sentence embedding. 
The early work resorts to word embedding~\cite{bengio2003neural,mikolov2013distributed,pennington2014glove} and represents an input sentence by a pooling vector (mean or weighted mean) over all embeddings of its words~\cite{kiros2015skip,wieting2015paraphrase}. More recently, sentence embedding obtained from pre-trained language models (PLMs) made a breakthrough thanks to PLM's powerful ability in modeling global context~\cite{peters-etal-2018-deep,devlin2019bert,liu2019roberta}, and it quickly became the standard practice for sentence embedding.  


Despite the success of sentence embedding from PLMs, an embedded vector (Sen2Vec) representing a point estimate does not naturally express uncertainty about the target concepts associated with the input sentence \cite{vilnis2015word}. 
In essence, this uncertainty originates from many-to-one nature in language representation: (1) \textbf{model uncertainty}: model uncertainty refers to the notion of randomness caused by inherently random effects within the model. (i.e. one sentence may have many representations according to different representations within the same model (e.g., dropout)) Considering that these representation are from the same sentence, they should remain close to each other; (2) \textbf{data uncertainty}: many sentences with different linguistic structure (e.g., paraphrase) may have the same meaning in semantics. Considering the identical semantics, their corresponding representation should get close with each other.

When quantifying uncertainty, we assume that close-semantic sentences' representation follows the same probability distribution. Given a sentence, since the model only observes one sample, it is natural to ask how much a language model can capture such a rich distribution.

A natural solution to this issue is to merge such a probabilistic perspective into sentence embedding, which represents a sentence as a distribution $P(\mu,\Sigma)$, where $\mu$ is the mean and covariance $\Sigma$ intuitively portrays the uncertainty of the distribution $P$. 
Unfortunately, there is a critical challenge to putting this idea into practice: previous works \cite{bamler2017dynamic,camacho2018word,zhou2019density} are only plausible on word embedding and require retraining word embedding with probabilistic embedding on large-scale data to advance SOTA. It is costly even for training a PLM without probabilistic embedding~\cite{devlin2019bert,radford2019language,he2020deberta,clark2019electra,raffel2020exploring}, which typically consumes considerable GPU computations for weeks. 

In this paper, we propose an efficient framework for probabilistic sentence embedding (Sen2Pro) from PLMs that represents a sentence as a probability density in an embedding space to reflect the uncertainty in the sentence representation.


Concerning \textbf{model uncertainty} and \textbf{data uncertainty}, we propose two simple methods to quantify both on top of a PLM. Specifically, to measure model uncertainty, we assume a sentence vector is drawn from a distribution $P(\mu^m, \Sigma^m)$, which can be estimated by many representations of the targeted sentence using a set of stochastic PLMs obtained from Monte Carlo dropout \cite{gal2016dropout}. 

Similarly, we apply the data augmentation technique to generate many semantically equivalent sentences to the targeted sentence for measuring data uncertainty. Then we assume a sentence vector is drawn from a distribution $P(\mu^d, \Sigma^d)$, which the representations of the augmented sentences from the PLM can estimate. 
In addition, we also introduce some ways to utilize both $\mu^{m}$ ($\mu^d$) and $\Sigma^m$ ($\Sigma^d$) as the final sentence representation for different downstream tasks.

Moreover, drawing from previous works \cite{chen2016infogan,li2019rate,gao2019rate,tschannen2019mutual,grohs2021phase} that explored the relationships between deep learning representation and relative entropy, we present theoretical explanations of why our probabilistic sentence embedding (Sen2Pro) is superior to point vector-based sentence embedding (Sen2Vec). Meanwhile, extensive experiments demonstrate the practical effectiveness of Sen2Pro on text classification, semantic similarity match, dialogue generation evaluation, and machine translation evaluation. Besides, Sen2Pro demonstrates its superiority in capturing sentence-level linguistic analogy over Sen2Vec.

\section{Related Work}
\subsection{Sentence Embedding}
Methods for sentence embedding learning have been extensively explored, and all these methods represent a sentence as a point embedding. 
Early works use the weighted sum of word embedding to represent a sentence. 
Then some methods based on the distributional hypothesis have been done. Hill \cite{hill2016learning} learned sentence representations with the internal structure of each sentence, and Kiros \cite{kiros2015skip} followed the idea of Word2Vec \cite{mikolov2013distributed} to represent a sentence by predicting its surrounding sentences. In recent years, the pre-trained language model \cite{devlin2019bert} has become the standard sentence paradigm because of its strong ability to capture syntactic and semantic features of sentences by learning from the large-scale corpus. 
Furthermore, several researchers used contrastive learning to augment sentence representation \cite{zhang2020unsupervised,yan2021consert,meng2021coco,gao2021simcse,wang2021cline}, based on the assumption that a high-quality representation method should bring similar sentences closer while pushing away dissimilar ones. 
They belong to Sen2Vec and thus fail to model uncertainty. This paper goes beyond point sentence embedding and explores probabilistic sentence embedding.

\subsection{Probabilistic Word Embedding}
In NLP, probabilistic embedding originated from word embedding \cite{bengio2003neural,mikolov2013distributed}, and existing probabilistic embedding methods work only on words, where a word from the vocabulary is represented as a density distribution in an embedding space. 
Although variants of probabilistic word embedding have been developed \cite{vilnis2015word,bamler2017dynamic,camacho2018word,zhou2019density}, they used a similar paradigm, which adapts the Skip-gram model \cite{mikolov2013distributed} with a non-parametric approach. 
Specifically, a Skip-gram model is retrained as the density distribution with a specific word sampling (e.g., synonym) method and a specific loss function (e.g., a margin loss). 
Therefore, existing probabilistic embedding needs an extremely time-consuming retraining stage and can not be applied to PLMs (e.g., BERT). 
Different from them, this paper contributes to the literature by developing probabilistic embedding for sentences that serves as a plug-and-play method on pre-trained language models \cite{devlin2019bert} without any time-consuming retraining stage.
 
\section{Methodology: Sen2Pro}
Because of the many-to-one nature of sentence embedding, we model the uncertainty from two perspectives, i.e., model uncertainty and data uncertainty. Accordingly, we assume that the representation of one sentence follows a distribution $P(\mu,\Sigma)$, which measures either model uncertainty or data uncertainty. The goal of our Sen2Pro framework is to estimate the two distributions $P(\mu,\Sigma)$ for a sentence embedding based on a pre-trained language model $f_{\theta}$ (e.g., BERT), where $\theta$ denotes its parameters. 
In general, there are two steps in Sen2Pro: the sampling stage and the estimation stage. 
The sampling stage generates embedding instances to capture two kinds of uncertainties: model uncertainty and data uncertainty (\S \ref{sec3p}). 
The estimation stage aims to estimate the parameters in the density distributions (i.e., mean vector and covariance matrix) based on the embedding instances (\S \ref{sec43}). 
After both distributions are estimated, the general idea to apply them to specific tasks is presented (\S \ref{use}). 

\subsection{Sampling Stage}

\paragraph{Model Uncertainty}\label{sec3p}
Model uncertainty originates from the fact that one sentence may have different representations due to inherent randomness within models.
In this paper, we use a pre-trained language model $f_{\theta}$ and try to quantify \emph{model uncertainty}. Considering that the key ingredient of model uncertainty is to vary the model while keeping the sentence $s$ unchanged, we utilize MC dropout to create embedding instances for quantifying model uncertainty.
Specifically, for each sentence ${s}$, we utilize MC Dropout \cite{gal2016dropout,lakshminarayanan2017simple} over the parameters $\theta$ as sampling, and repeat sampling $N$ times to obtain different subsets of the parameters $\theta$: $\{\widehat{\theta}_{i}\mid i=1, \ldots, N\}$. 
In this way, we generate a set of embeddings as follows:
\begin{equation}\label{eq000}
    \mathcal{S}^m = \big\{x_i=f_{\widehat{\theta}_{i}}\left( s\right)\mid i=1, \ldots, N\big\}
\end{equation}
As shown in Eq.~\ref{eq000}, each subset of model's parameters represent a sub-structure of the model, which naturally matches with the definition of model uncertainty.

\paragraph{Data Uncertainty}
Data uncertainty corresponds to the many-to-one nature of sentence embedding. In other words, sentences that are semantically similar but slightly different should have close representations. 
Data uncertainty exists in universal real-world scenarios: since the model requires lots of training data to perform well, it is common to augment high-quality labeled sentences with lower-quality web-crawled data to save time and effort. 
To naturally imitate such uncertainty, in this paper, a simple data augmentation method, word-level operation, is applied to the input sentence, which adds proper noise to an input sentence.
After repeating data augmentation $N$ times (i.e., randomly dropping a word in ${s}$, swapping two words, replacing or inserting a word in ${s}$ with any word from vocabulary) for the input sentence $ s$, a set of new sentences, $ s_{1}, s_{2},\ldots, s_{N}$ are obtained. 
Then, the sentences are fed to the pre-trained model $f_{\theta}$ to get $N$ embeddings. In this way, we can obtain a set of embeddings as follows:

\begin{equation}
    \mathcal{S}^d = \big\{x_i=f_{\theta}\left( s_i\right)\mid i=1, \ldots, N\big\}
\end{equation}

\subsection{Estimation Stage}\label{sec43}
After obtaining the required embedding instances for model and data uncertainty, we can estimate the probability distributions on them, respectively. Similar to cases of \cite{kendall2017uncertainties,maddox2019simple,ovadia2019can,abdar2021review,hullermeier2021aleatoric}, we do estimation towards two uncertainty individually rather than unifying, which is also empirically verified in Appendix~\ref{compare}.
Suppose $\mathcal{S}$ denotes either $\mathcal{S}^m$ or $\mathcal{S}^d$, it is natural to estimate its mean and covariance as follows:
\begin{equation}
    \mu= \frac{1}{|\mathcal{S}|} \sum_{x\in \mathcal{S}} x
\end{equation}
\begin{equation}
        \Sigma = \frac{1}{|\mathcal{S}|} \sum_{x\in \mathcal{S}} (x-\mu)(x-\mu)^\top
        \label{eq:cov}
\end{equation}
where $|\mathcal{S}|$ means the size of the set $\mathcal{S}$ and $(.)^{\top}$ is an transpose operation. 
We use $\mu^m$ and $\Sigma^m$ respectively to denote the statistics estimated from model uncertainty (i.e., $\mathcal{S} = \mathcal{S}_m$), and $\mu^d$ and $\Sigma^d$ denote those estimated through data uncertainty (i.e., $\mathcal{S} = \mathcal{S}_d$). 

However, such a simple covariance matrix estimator (SCE) owns severe problems on both theoretical \cite{xiao2012covariance} and practical sides: it is known to degrade rapidly as an estimator as the number of variables increases, thus performing badly in a high-dimensional case (e.g., 768 dimensions in BERT).
To address this issue, inspired by \citet{bien2016convex}, we instead employ the banding estimator, which uses an off-diagonal entry removal operation on the covariance matrix.

Specifically, for a covariance matrix $\Sigma=\left(\Sigma_{i j}\right)_{k \times k}$ where $k$ is the dimension of $\Sigma$, we use $B(\Sigma)$ as the estimation of $\Sigma$ as follows:
\begin{equation}
    \hat \Sigma = B(\Sigma)=\operatorname{Diag}(\Sigma)
\end{equation}

Besides, Theorem~\ref{th1} provides an estimation error bound for our banding estimator, whose proof is presented in Appendix~\ref{theorem1}.
\begin{theorem}\label{th1}
Suppose $\Sigma$ is the covariance matrix of the ground truth distribution $P(\mu,\Sigma)$, and $\hat{\Sigma}$ denote $B(\Sigma)$, then we have
\begin{equation}
    \left\|\hat{\Sigma}-{\Sigma}\right\|_{2}=O_p\left(\left(\frac{\log k}{n}\right)^{M}\right)
\end{equation}
where $k$ is the dimension of $\Sigma$, and $M$ is a positive constant satisfying $M<\frac{1}{2}$. $O_p$ and $n$ means $\frac{\log k}{n})^{M}$ is stochastically bounded as $n \to \infty$.
\end{theorem}
Besides, SCE also in practicability problems compared to the banding estimator. 
SCE owns a significantly worse performance-efficiency trade-off than our banding estimator, which will be empirically verified in Sec~\ref{band}. Moreover, theoretical analyses for comparison between Sen2Vec and Sen2Pro are deferred to Appendix~\ref{theory}. 

\subsection{Usage of Sen2Pro}\label{use}
Unlike previous works on probabilistic embedding that drop $\hat \Sigma$ in tasks, in Sen2Pro, both mean vector $\mu$ (i.e., $\mu^m$ and $\mu^d$) and covariance vector $\hat \Sigma$ (i.e., $\hat \Sigma^m$ and $\hat \Sigma^d$) are used for sentence embedding. In the next section, we illustrate our strategies to use $\mu$ and $\hat \Sigma$, more details are presented in Sec~\ref{setting}.


\begin{table*}[!h]\small
\centering
\resizebox{\linewidth}{!}{
\begin{tabular}{@{}c|c|c|c|c|c|c|c|c|c@{}}
\toprule
Dataset                 & Model       & 10        & 200       & Full      & Dataset                  & Model       & 10        & 200        & Full    \\ \midrule
\multirow{2}{*}{AGNews} & BERT-base   & 69.5      & 87.5      & 95.2      & \multirow{2}{*}{DBPedia} & BERT-base   & 95.2      & 98.5      & 99.3    \\
                        & BERT-base-G & \color[HTML]{3166FF}74.4(1.1) & \color[HTML]{3166FF}90.2(0.3) & \color[HTML]{3166FF}95.6(0.1) &                          & BERT-base-G & \color[HTML]{3166FF}96.5(0.2) & \color[HTML]{3166FF}99.1(0.1) & 99.3(*) \\ \midrule
\multirow{2}{*}{Yahoo}  & BERT-base   & 56.2      & 69.3      & 77.6      & \multirow{2}{*}{IMDB}    & BERT-base   & 67.5      & 86.9      & 95.6    \\
                        & BERT-base-G & \color[HTML]{3166FF}60.5(1.8) & \color[HTML]{3166FF}72.9(0.6) & \color[HTML]{3166FF}78.2(0.2) &                          & BERT-base-G & \color[HTML]{3166FF}70.4(0.6) & \color[HTML]{3166FF}88.5(0.3) & \color[HTML]{3166FF}95.7(*) \\ \bottomrule
\end{tabular}}
\caption{Test accuracy(\%) comparison between Sen2Pro and Sen2Vec for text classification. The results on each dataset are the mean of three runs; the standard derivation (i.e., the values in brackets) is given for PLM-G, where * means the derivation is smaller than 0.1\%.}
\label{tab:2}
\end{table*}

\section{Experiment}\label{Experiment}

\begin{table*}[!h]\small
\centering
\begin{tabular}{@{}c|ccccc|c@{}}
\toprule
Baseline     & STS-12  & STS-13    & STS-14 & STS-15 &STS-16 & Avg   \\ \midrule
BERT& 57.86$\rightarrow$\color[HTML]{3166FF}59.55 & 61.97$\rightarrow$\color[HTML]{3166FF}66.20 & 62.49$\rightarrow$\color[HTML]{3166FF}65.19 & 70.96$\rightarrow$\color[HTML]{3166FF}73.50  & 69.76$\rightarrow$\color[HTML]{3166FF}72.10 & 63.69$\rightarrow$\color[HTML]{3166FF}66.70(+3.01) \\
BERT$_{l}$ & 57.74$\rightarrow$\color[HTML]{3166FF}59.90 & 61.16$\rightarrow$\color[HTML]{3166FF}66.20 & 61.18$\rightarrow$\color[HTML]{3166FF}65.62  & 68.06$\rightarrow$\color[HTML]{3166FF}73.01 & 70.30$\rightarrow$\color[HTML]{3166FF}74.72 & 62.62$\rightarrow$\color[HTML]{3166FF}67.47(+4.85) \\ \midrule
W-BERT      & 63.62$\rightarrow$\color[HTML]{3166FF}64.50 & 73.02$\rightarrow$\color[HTML]{3166FF}73.69 & 69.23$\rightarrow$\color[HTML]{3166FF}69.69 & 74.52$\rightarrow$\color[HTML]{3166FF}74.69 &  72.15$\rightarrow$\color[HTML]{3166FF}76.11 & 69.21$\rightarrow$\color[HTML]{3166FF}70.39 (+1.18)  \\ 

W-BERT      & 64.02$\rightarrow$\color[HTML]{3166FF}64.90 & 73.27$\rightarrow$\color[HTML]{3166FF}73.94 & 69.58$\rightarrow$\color[HTML]{3166FF}70.04 & 74.77$\rightarrow$\color[HTML]{3166FF}74.94 &  72.50$\rightarrow$\color[HTML]{3166FF}76.44 & 69.58$\rightarrow$\color[HTML]{3166FF}70.69 (+1.26)  \\ 

C-BERT      & 64.09$\rightarrow$\color[HTML]{3166FF}65.01 & 78.21$\rightarrow$\color[HTML]{3166FF}78.54 & 68.68$\rightarrow$\color[HTML]{3166FF}69.04 & 79.56$\rightarrow$\color[HTML]{3166FF}79.90 & 75.41$\rightarrow$\color[HTML]{3166FF}75.74 & 72.27$\rightarrow$\color[HTML]{3166FF}72.69 (+0.42)  \\ 
C-BERT$_{l}$      & 70.23$\rightarrow$\color[HTML]{3166FF}70.70 & 82.13$\rightarrow$\color[HTML]{3166FF}82.54 & 73.60$\rightarrow$\color[HTML]{3166FF}74.12 & 81.72$\rightarrow$\color[HTML]{3166FF}82.01 & 77.01$\rightarrow$\color[HTML]{3166FF}77.58 & 76.03$\rightarrow$\color[HTML]{3166FF}76.48 (+0.45)  \\ \midrule
Sim-BERT      & 68.93$\rightarrow$\color[HTML]{3166FF}69.33 & 78.68$\rightarrow$\color[HTML]{3166FF}78.93 & 73.57$\rightarrow$\color[HTML]{3166FF}73.95 & 79.68$\rightarrow$\color[HTML]{3166FF}80.01 & 79.11$\rightarrow$\color[HTML]{3166FF}79.29 & 75.11$\rightarrow$\color[HTML]{3166FF}75.44 (+0.33)  \\ 
Sim-BERT$_{l}$      & 69.25$\rightarrow$\color[HTML]{3166FF}69.60 & 78.96$\rightarrow$\color[HTML]{3166FF}79.30 & 73.64$\rightarrow$\color[HTML]{3166FF}73.92 & 80.06$\rightarrow$\color[HTML]{3166FF}80.31 & 79.08$\rightarrow$\color[HTML]{3166FF}79.42 & 75.31$\rightarrow$\color[HTML]{3166FF}75.61 (+0.30)  \\ 
\bottomrule
\end{tabular}
\caption{The experimental results of adding our Sen2Pro on widely used sentence embedding methods. Specifically, BERT$_{l}$ means BERT$_{large}$.}
\label{table:2}
\end{table*}

\begin{table}[!h]\small\centering
\begin{tabular}{@{}c|ccc@{}}
\toprule
Sentence Embedding  & MRR            & Hits@1                                                                    & Hits@3          \\ \midrule
BERT                & 68.01          & 51.70                                                                    & 81.91          \\
BERT+Ours           & \color[HTML]{3166FF}{68.69} & \color[HTML]{3166FF}{52.24}                                                           & \color[HTML]{3166FF}{82.63} \\ \midrule
BERT-whitening      & 66.58          & 46.54                                                                    & 84.22          \\
BERT-whitening+Ours & \color[HTML]{3166FF}{67.49} & \color[HTML]{3166FF}{48.23}                                                           & \color[HTML]{3166FF}{84.56} \\ \midrule
Sentence-BERT       & 64.12          & 47.07 & 79.05          \\
Sentence-BERT+Ours  & \color[HTML]{3166FF}{66.10} & \color[HTML]{3166FF}{48.55}                                                           & \color[HTML]{3166FF}{80.34} \\ \midrule
SimCSE              & 69.50          & 52.34                                                                    & 84.43          \\
SimCSE+Ours         & \color[HTML]{3166FF}{70.01} & \color[HTML]{3166FF}{52.68}                                                           & \color[HTML]{3166FF}{84.69} \\ \bottomrule
\end{tabular}
\caption{Performances on EvalRank \cite{wang2022just} using Sen2Vec and Sen2Pro. `+Ours' means Sen2Pro, and the results are the mean of five runs to show the statistical significance with $p$ value \textless 0.05.}
\label{tab1}
\end{table}

This section comprehensively evaluates the effectiveness of our Sen2Pro framework on the following various tasks. The results consistently show that Sen2Pro outperforms Sen2Vec. Specifically, we choose two sets of tasks for evaluation: Fine-tune-need Task and Fine-tune-free Task.

\subsection{Basic Setting}\label{setting}
Recall that $\mu^m$ and $\hat \Sigma^m$ ($\mu^d$ and $\hat \Sigma^d$) denote the estimated mean and covariance from model (data) uncertainty. Then we present how to use them for different tasks. 
\paragraph{Fine-tune-need Task: Text Classification}\label{usage}
After estimation stage in data and model uncertainty,
each sentence is represented as the concatenation of $\frac{\mu^m + \mu^d}{2}$ and (diagonal entries of) $\frac{\hat \Sigma^m + \hat \Sigma^d}{2}$. Specifically, we use reparameterization to handle the non-differentiable issue of the sampling process.

\paragraph{Fine-tune-free Task: Sentence Similarity, Dialogue Evaluation, Neural Machine Translation Evaluation} 
For such NLP tasks, the distance between two representations is needed. Although KL divergence is natural to measure the distance between two distributions, it has practical drawbacks. For example, in BERT-base case where $\mu \in \mathcal{R}^{768 \times 1}$, $\hat\Sigma \in \mathcal{R}^{768 \times 768}$, KL divergence not only is time-consuming but also results in numerical errors, considering that KL divergence includes operations $det(\hat\Sigma)$ and $\hat\Sigma^{-1}$. In practice, most entries of $\hat\Sigma$ are between 0 and 0.5, so the determinant becomes extremely small, which leads to numerical errors. Moreover, the computation of $\hat\Sigma^{-1}$ will be unstable when the dimension is high. 

Therefore, a simple function is taken to measure the distance between two probabilistic sentence embeddings $(\mu_{a},\Sigma_{a})$ and $(\mu_{b},\Sigma_{b})$ as follows:
\begin{align}\label{eq2}
     &d(\mathcal{N}(\mu_{a},\Sigma_{a}),\mathcal{N}(\mu_{b},\Sigma_{b}))\\ \nonumber
     &= (1-\alpha)l_{1}(\mu_{a}-\mu_{b}) + \alpha l_{1}(\Sigma_{a}-\Sigma_{b}) \nonumber
\end{align}
where $l_{1}$ represents the $l_{1}$-norm and $\alpha$ is to balance the two terms. For $\alpha$, we consider it as a balance factor for different magnitudes of $\mu$ and $\Sigma$, defined as follows:
\begin{equation}
    \alpha = \frac{l_{1}(\mu_{a}-\mu_{b})}{l_{1}(\Sigma_{a}-\Sigma_{b})}
\end{equation}
In most cases, $\alpha$ ranges from 0.01 to 0.05. Besides, we will try to apply Sen2Pro on other evaluation tasks \cite{shen2022revisiting,shen2022evaluation} like paraphrase, data-to-text, and summarization.

\subsection{Text Classification}
\paragraph{Benchmarks and Setting}
We choose four widely-used text classification benchmarks: AG News \cite{zhang2015character}, DBpedia \cite{maas2011learning}, Yahoo! Answers \cite{chang2008importance} and IMDB \cite{maas2011learning}. The evaluation metric is the test accuracy, and the best performance is selected based on validation set accuracy. The baseline is Sen2Vec, with 15 data augmentation samples per sentence. In Sen2Pro, we choose the BERT-base model as the PLM and use the `first-last-avg'(a pooling way that average first and last layer representation). The sampling number is 15 for the model uncertainty and data uncertainty.
Moreover, we use two settings for model evaluation: few-shot and full-dataset. In the few-shot setting, models are trained with randomly selected 10 and 200 labeled sentences per class. In the full-dataset setting, models are trained with the whole training set. 
\paragraph{Performances}
The results are listed in Table~\ref{tab:2}. Our Sen2Pro consistently performs better than Sen2Vec under few-shot settings because Sen2Pro contains more semantic information about a sentence due to its probabilistic characteristic. Moreover, Sen2Pro achieves comparable or better performances in the full training setting than Sen2Vec.

\subsection{Sentence Similarity}
\paragraph{Benchmarks and Setting}
We use seven commonly-used STS datasets \cite{agirre2012semeval,agirre2013sem,agirre2014semeval,agirre2015semeval,agirre2016semeval} for evaluation. Besides, considering the limitation of traditional intrinsic evaluation \cite{wang2022just}, we choose EvalRank \cite{wang2022just} for linguistic analogy evaluation, which overcomes the previous limitation. In Sen2Pro, we use several state-of-the-art pre-trained language models, including BERT-base \cite{devlin2019bert}, BERT-whitening \cite{su2021whitening,huang2021whiteningbert}, Sentence-BERT \cite{reimers2019sentence}, and SimCSE \cite{gao2021simcse}. Specifically,  EvalRank uses the mean reciprocal rank (MRR) and Hits@k scores for evaluation, and a higher score indicates a better embedding model. The sampling number for each uncertainty is set as 15.


\paragraph{Performances}
The results of Sen2Pro are reported in Table \ref{table:2} and ~\ref{tab1}, which illustrate that Sen2Pro outperforms Sen2Vec with a substantial gap under all settings. Moreover, Sen2Pro using the`base' PLMs can achieve better or comparable performance than Sen2Vec using `large' PLMs.


\begin{table}[!h]
\centering
\resizebox{\linewidth}{!}{
\begin{tabular}{@{}ccccccc@{}}
\toprule
\multirow{2}{*}{Metric} & \multicolumn{2}{c}{Daily(H)}  & \multicolumn{2}{c}{Convai2}   & \multicolumn{2}{c}{Empathetic}    \\ \cmidrule(l){2-7} 
                        & Spr           & Pr            & Spr           & Pr            & Spr           & Pr            \\ \midrule
BLEU                    & 44.5          & 44.4          & 80.0          & 80.1          & 13.6          & 33.1          \\
METEOR                  & 1.8          & 5.0          & 80.0          & 76.7          & 38.2          & 13.3          \\
ROUGE-L                 & 54.5          & 41.7          & 20.0          & 6.1          & 39.1          & 47.2          \\
Greedy                  & 85.5          & 76.4          & 60.0          & 79.4          & 73.6          & \color[HTML]{3166FF}86.4          \\
Average                 & 20.9          & 20.9          & 60.0          & 87.9          & 66.4          & 72.5          \\
Extrema                 & 74.5          & 76.1          & 80.0          & 76.6          & 61.8          & 72.2          \\
B-SCORE                 & 85.5          & \color[HTML]{3166FF}85.7          & 80.0          & \color[HTML]{3166FF}93.9          & 60.0          & 69.7          \\
Sen2Vec                 & 69.9          & 76.1          & 60.0          & 63.9          & 55.2          & 60.3          \\
Sen2Pro                 & \color[HTML]{3166FF}{87.3}          &81.4  &\color[HTML]{3166FF}{100}  &{85.9}  &\color[HTML]{3166FF}{80.9}  & {81.6} 
\\\bottomrule
\end{tabular}}
\caption{Correlations of the evaluation metrics on the dialogue corpora. `Spr.' and `Pr.' refer to Spearman and Pearson correlation coefficients, respectively; B-SCORE represents BERTScore.}
\label{tab0}
\end{table}

\begin{table*}[!h]\small
\centering
\begin{tabular}{@{}c|ccccccc|c@{}}
\toprule
Metric    & cs-en                        & de-en                        & fi-en                        & lv-en                        & ru-en                        & tr-en                        & zh-en                        & Avg                          \\ \midrule
BLEU      & 97.1                        & 92.3                        & 90.3                        & 97.9                        & 91.2                        & 97.6                        & 86.4                        & 93.3                        \\
CDER      & 98.9                        & 93.0                        & 92.7                        & 98.5                        & 92.2                        & 97.3                        & 90.4                        & 94.7                        \\
BLEND     & 96.8                        & 97.6                        & 95.8                        & 97.9                        & { 96.4} & 98.4                        & 89.4                        & 96.0                        \\
BERTScore & 99.6                        & {\color[HTML]{3166FF} 98.2} & 94.7                        & 97.9                        & 956                        & 98.6                        & {\color[HTML]{3166FF} 98.4} & 97.6                        \\
Sen2Vec   & 97.3                        & 92.5                        & 93.1                        & 98.6                        & 90.1                        & 98.0                        & 92.1                        & 94.5                        \\
Sen2Pro   & {\color[HTML]{3166FF} 99.8} & 96.2                        & {\color[HTML]{3166FF} 99.0} & {\color[HTML]{3166FF} 99.5} & \color[HTML]{3166FF} 96.5                        & {\color[HTML]{3166FF} 99.0} & 97.9                        & {\color[HTML]{3166FF} 97.8} \\ \bottomrule
\end{tabular}
\caption{Pearson correlations with system-level machine translation evaluation on WMT17.}
\label{tabnmt1}
\end{table*}

\subsection{Dialogue Evaluation}
\paragraph{Benchmark and Setting}
We choose three widely-used dialogue benchmarks: Daily(H) \cite{lowe2017towards}, Convai2 \cite{dinan2020second}, and Empathetic \cite{rashkin2019towards}. Each benchmark consists of dialogue queries, the corresponding responses, and human-annotated responses. For baseline metrics, we choose BLEU \cite{papineni2002bleu}, ROUGE \cite{lin2004rouge}, METEOR \cite{denkowski2014meteor}, Greedy Matching \cite{rus2012optimal}, Embedding Average \cite{wieting2015towards}, Vector Extrema \cite{forgues2014bootstrapping} and BERTScore \cite{zhang2019bertscore}. For Sen2Pro, we use the BERT-base model and the `first-last-avg' representation and set the sampling number of uncertainty to 15.
\paragraph{Performances}
The performances of various evaluation metrics are reported in Table~\ref{tab0}. Sen2Pro performs best in most cases, demonstrating its good generalization ability and robustness for the dialogue evaluation task.

\begin{table*}[!h]\small
\centering
\begin{tabular}{@{}c|ccccccc|c@{}}
\toprule
Model            & STS-12 & STS-13 & STS-14 & STS-15 & STS-16 & STS-B & SICK-R & Average          \\ \midrule
BERT-base        & 57.84  & 61.95  & 62.48  & 70.95  & 69.81  & 59.04 & 63.75  & 63.69        \\
BERT-base-$G_{m+d}$    & \color[HTML]{3166FF}59.40  & \color[HTML]{3166FF}63.03  & \color[HTML]{3166FF}64.18  & \color[HTML]{3166FF}71.97  & \color[HTML]{3166FF}70.73  & \color[HTML]{3166FF}62.59 & \color[HTML]{3166FF}64.69  & \color[HTML]{3166FF}65.23 (+1.54) \\
BERT-base-$G_{m}$      & 58.86  & 62.66  & 63.77  & 71.48  & 70.56  & 61.89 & 64.28  & 64.79 (+1.10) \\
BERT-base-$G_{d}$ & 58.14  & 62.20  & 62.83  & 71.00  & 70.36  & 59.30 & 64.10  & 63.99 (+0.30) \\\bottomrule
\end{tabular}
\caption{Ablation studies on the two uncertainties in STS. BERT-base-$G_{m}$ and BERT-base-$G_{d}$ represent Sen2Pro with only model uncertainty or data uncertainty, respectively.}
\label{tab3}
\end{table*}

\subsection{NMT Evaluation}
\paragraph{Benchmark and Setting}
We work on the WMT-17 machine translation benchmark \cite{bojar2017results}. Moreover, BLEU \cite{papineni2002bleu}, CDER \cite{leusch2006cder}, BLEND \cite{ma2017blend}, Sen2Vec, and BERTScore \cite{zhang2019bertscore} are chosen as baseline metrics. The setting of Sen2Pro is the same as the one of dialogue evaluation. 
\paragraph{Performances}
The results are shown in Table~\ref{tabnmt1}. Our Sen2Pro achieves comparable performance towards a top metric (i.e., BERTScore) and significantly outperforms Sen2Vec, which demonstrates the effectiveness of Sen2Pro in the NMT evaluation task. Segment-level results are shown in Appendix~\ref{segment}.
The results indicate that our Sen2Pro can yield competitive performance to SOTA as an automatic metric to evaluate dialogue generation.


\section{Analysis and Discussion}
This section presents analyses of Sen2Pro, and the BERT model is used as the PLM in the analyses.
Specifically, the representation uncertainty analysis is made on the STS task, and detailed results are deferred to Appendix~\ref{importance} and \ref{performance}.
Moreover, we choose the BERT model as our pre-trained sentence encoder in the analysis.

\begin{figure}[!h]
    \centering
    \includegraphics[scale=0.45]{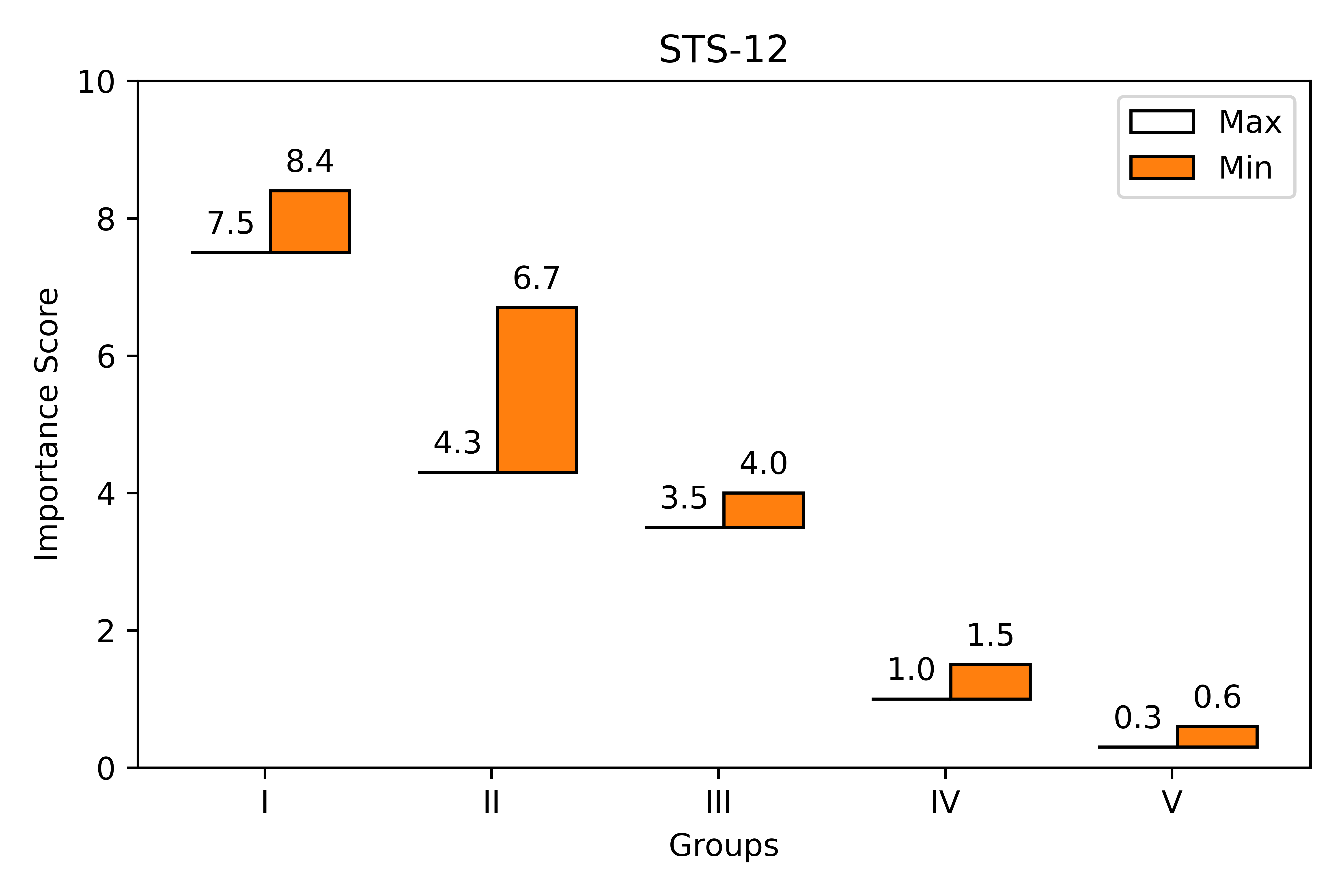}
    \caption{The importance score of five groups of features. `\uppercase\expandafter{\romannumeral1}' represents the features with top 20\% highest uncertainty; `\uppercase\expandafter{\romannumeral5}' indicates the features with bottom 20\% lowest uncertainty; `Min' and `Max' mean the lowest and highest importance score under three runs.}
    \label{fig:1}
\end{figure}

\paragraph{Feature with higher model uncertainty is more important}
We investigate the relation between the model uncertainty and the \textbf{feature importance} on the STS task. In Sen2Pro, a sentence is represented as $\mu^{m} \in \mathcal{R}^{768 \times 1}$ and diagonal entries of $\hat \Sigma^{m} \in \mathcal{R}^{768 \times 768}$. From another perspective, a sentence is represented by 768 features in $\mu^{m}$, and $\hat \Sigma^{m}$ reflects the corresponding model uncertainties of such features. Let $T$ be the feature set, and $t \in T$ is a feature subset. We define the \textbf{feature importance} as follows:
\begin{equation}
    score(t) = |\rho(T)-\rho(T/t)|
\end{equation}
where $\rho$ represents Spearman's correlation in STS evaluation, and $score(t)$ describes the performance change after removing the feature subset $t$ from $T$. Then we separate the 768 features into five groups according to their uncertainty $\hat \Sigma^{m}$, and name them as `\uppercase\expandafter{\romannumeral1}' to `\uppercase\expandafter{\romannumeral5}', as shown in Figure~\ref{fig:1}. When a group of features is removed from $T$, these features are set to 0 in $\mu^{m}$ and $\hat \Sigma^{m}$, respectively. Figure~\ref{fig:1} lists the results on STS-12. In Figure~\ref{fig:1}, as the feature's uncertainty decreases, the importance score drops, indicating that features with higher uncertainty are more important in the STS task.

\begin{table*}[!h]\footnotesize
\centering
\resizebox{\linewidth}{!}{
\begin{tabular}{@{}c|c|c|c|c|c|c|c|c|c@{}}
\toprule
Dataset                 & Model       & 10        & 200       & Full      & Dataset                  & Model       & 10        & 200        & Full    \\ \midrule
\multirow{4}{*}{AGNews} & BERT-base   & 69.5      & 87.5      & 95.2      & \multirow{4}{*}{DBPedia} & BERT-base   & 95.2      & 98.5      & 99.3    \\
                        & BERT-base-G & \color[HTML]{3166FF}73.4(0.5) & \color[HTML]{3166FF}89.2(0.3) & 95.4(0.1) &                          & BERT-base-G & \color[HTML]{3166FF}96.3(0.2) & \color[HTML]{3166FF}99.1(0.1) & 99.3(*) \\
                        & BERT-base-$G_{m}$ & 71.6(1.0) & 88.3(0.4) & 95.2(0.2) &                          & BERT-base-$G_{m}$ & 95.8(0.2) & 98.8(*)     & 99.3(*) \\
                        & BERT-base-$G_{d}$ & 73.0(0.4) & 89.0(0.3) & 95.4(0.1) &                          & BERT-base-$G_{d}$ &  96.3(0.3) & 99.1(0.1) & 99.3(*) \\ \midrule
\multirow{4}{*}{Yahoo}  & BERT-base   & 56.2      & 69.3      & 77.6      & \multirow{4}{*}{IMDB}    & BERT-base   & 67.5      & 86.9      & 95.6    \\
                        & BERT-base-G & \color[HTML]{3166FF}60.1(0.8) & \color[HTML]{3166FF}72.4(0.6) & 78.0(0.1) &                          & BERT-base-G & \color[HTML]{3166FF}70.3(0.6) & \color[HTML]{3166FF}88.2(0.3) & 95.7(*) \\
                        & BERT-base-$G_{m}$ & 58.1(1.0) & 70.3(0.5) & 77.9(*)     &                          & BERT-base-$G_{m}$ & 69.2(0.6) & 87.5(0.2) & 95.6(*) \\
                        & BERT-base-$G_{d}$ & 59.9(0.7) & 71.7(0.3) & 78.0(0.1) &                          & BERT-base-$G_{d}$ & 69.8(0.4) & 88.2(0.1) & 95.7(*) \\ \bottomrule
\end{tabular}}
\caption{Ablation studies on the two uncertainties in text classification. }
\label{tab4}
\end{table*}

\begin{figure}[!htp]
    \centering
    \includegraphics[scale=0.45]{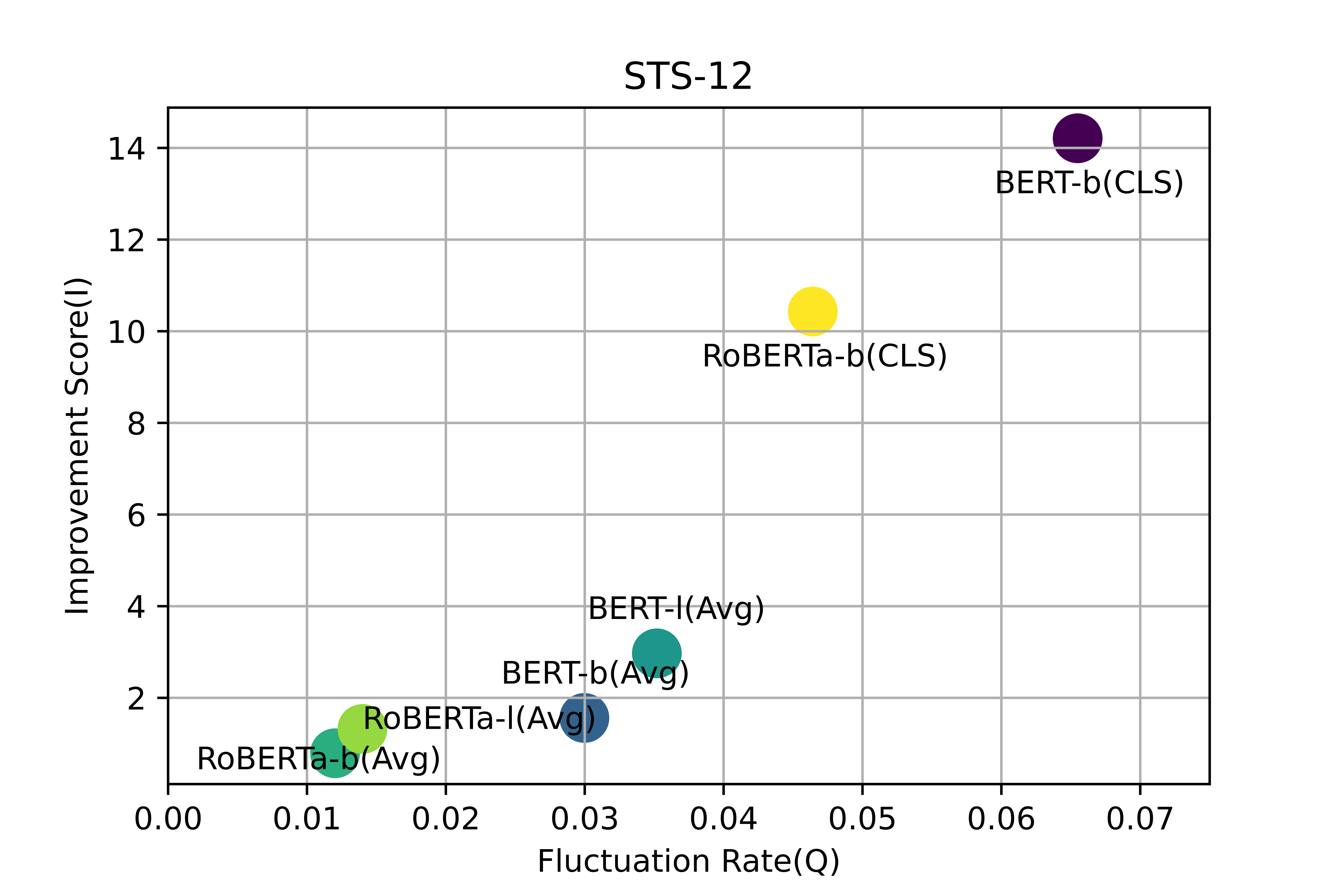}
    \caption{The relation between fluctuation rate ($Q$) and improvement score ($I$). `-b' and `-l' indicate `base' and `large', and `Avg' and `CLS' represent the `first-last-avg' and `[CLS]' representation.}
    \label{fig:2}
\end{figure}

\paragraph{Sen2Pro brings more benefits when model uncertainty are higher}
We also investigate the relation between the model uncertainty and the performance improvement of Sen2Pro over Sen2Vec on the STS task. Firstly, we define a metric called \textbf{Fluctuation Rate} $Q$ as follows:
\begin{equation}
  Q(f,D) = \frac{\Sigma_{i=1}^{|D|}(\Sigma_{j=1}^{k} \sigma_{ij})}{k \times |D|}
\end{equation}
where\footnote{Note that $\sigma$ corresponds to $\hat \Sigma^{m}$, we change the notation here to avoid repeating with the sum operation $\Sigma$.} $f$ and $D$ represent the PLM and benchmark, and $\sigma_{ij}$ is $j$-th element of the covariance matrix for $i$-th sentence in $D$, and $k$ denotes the dimension of model $f$. Such a metric can generally reflect the uncertainty of a specific model towards a specific benchmark. Based on $Q(f,D)$, we define the improvement score $I$ reflecting the improvement of Sen2Pro over Sen2Vec as follows:
\begin{equation}
    I = P_{Sen2Pro}(f,D) - P_{Sen2Vec}(f,D)
\end{equation}
where $P_{Sen2Pro}(f,D)$ and $P_{Sen2Vec}(f,D)$ represent the performance of model $f$ on $D$ under Sen2Pro and Sen2Vec, respectively. Besides, we add the result of [CLS] representation into experiments since the [CLS] representation owns a significantly higher fluctuation rate than the `first-last-avg' representation. The results on STS-12 are illustrated in Figure~\ref{fig:2}. As the fluctuation rate increases, the improvement score becomes more significant. Such empirical results demonstrate that the effectiveness of Sen2Pro is highly correlated to the model uncertainty and show Sen2Pro's superiority over Sen2Vec owns a positive correlation to model uncertainty.

\paragraph{Effectiveness of Two Uncertainties}\label{ablation}
We verify the effect of model and data uncertainty on Sen2Pro's performances. Specifically, Sen2Pro is evaluated on one intrinsic evaluation (STS) and one downstream task (text classification), and the results are demonstrated in Table~\ref{tab3} and Table~\ref{tab4}, respectively. As shown in Table~\ref{tab3}, the performance decreases when the data uncertainty is applied alone for the STS task since sentences' semantics may be changed due to the data augmentation. In contrast, the model uncertainty consistently brings benefits to the sentence representation. For text classification, both uncertainties improve the representation's generalization. Specifically, the contribution of the data uncertainty is higher than the model uncertainty. These empirical results also illustrate the usage of Sen2Pro (\S \ref{use}).

\begin{figure}[!htp]
    \centering
    \includegraphics[scale=0.45]{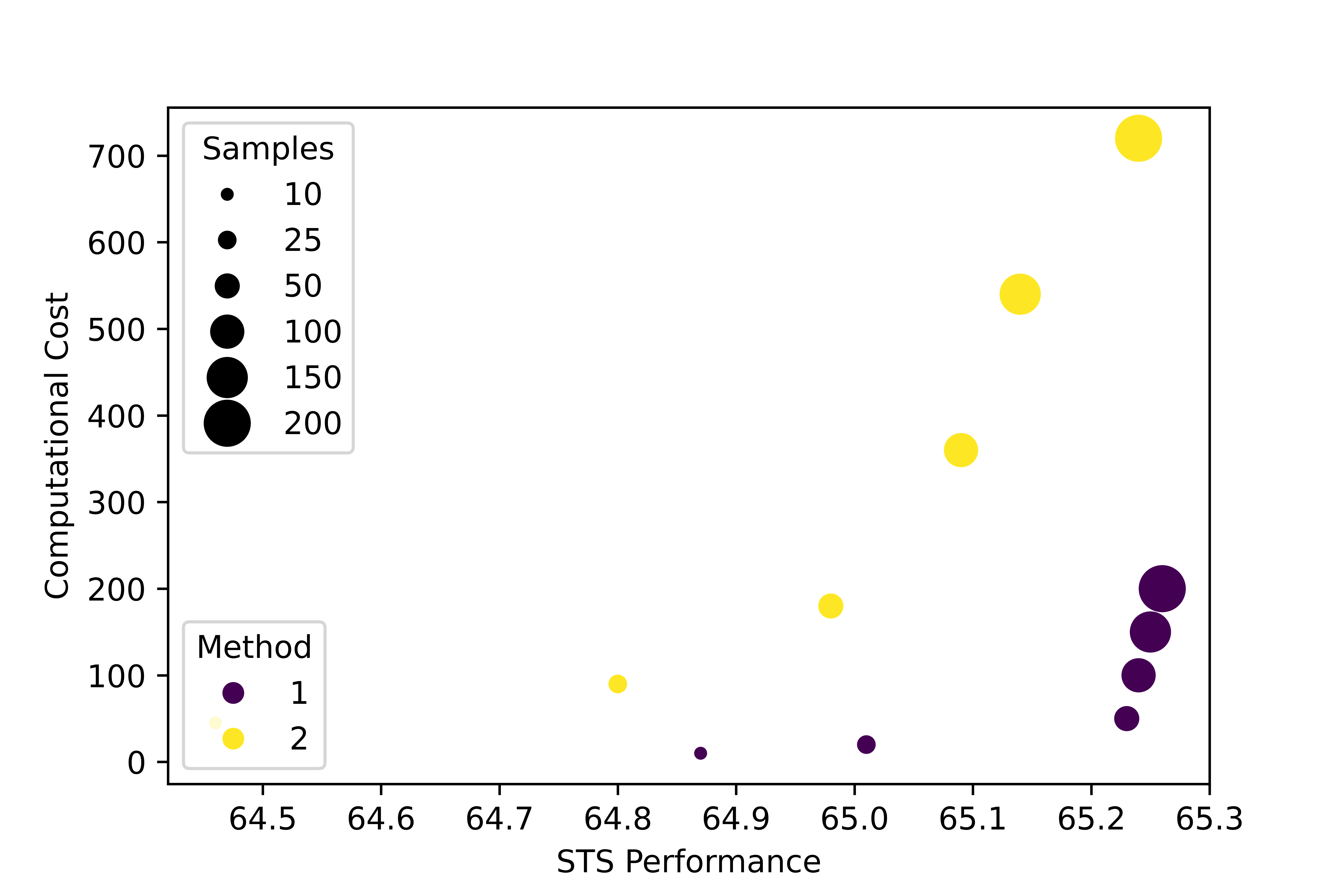}
    \caption{The performance-efficiency trade-off between SCE and our banding estimator, where `1' represents our estimation and `2' represents `SCE'.}
    \label{fig:00}
\end{figure}

\begin{table}[!h]
\centering
\begin{tabular}{@{}c|cccc@{}}
\toprule
Method            & STS & TC & Dialog & NMT        \\ \midrule
SCE        & 64.98  & 75.11  & 87.5  & 97.0  \\
Ours        &  \color[HTML]{3166FF}65.23  &  \color[HTML]{3166FF}75.45  &  \color[HTML]{3166FF}89.4  &  \color[HTML]{3166FF}97.8 \\\bottomrule
\end{tabular}
\caption{Comparisons between SCE and banding estimator. The number is the average performance. }
\label{tab000}
\end{table}
\paragraph{Effect of the Banding Estimator}\label{band}
As mentioned in Sec~\ref{sec43}, we use the banding estimator for covariance estimation. This part compares SCE (the usual estimator) and the banding estimator in two aspects: performance and efficiency. We choose BERT-base as the PLM, and the results are presented in Table~\ref{tab000}, and the efficiency is shown in Figure~\ref{fig:00}. As shown in Figure~\ref{fig:00}, our estimator achieves a significantly better performance-efficiency trade-off than SCE, demonstrating the banding estimator's effectiveness.


\section{Linguistic Case Study}
We conduct case studies following the famous analogy from Word2Vec \cite{mikolov2013distributed}. The widely used analogy takes the following form: Knowing that $A$ is to $B$ as $C$ is to $D$, given $l_{2}$ normalized embeddings $\vec{v}_A, \vec{v}_B, \vec{v}_C, \vec{v}_D$ for sentences $A, B, C, D$ for an analogy of the above form, the task compare the embedding distance $dis(A, B)$ and $dis(C, D)$, which is defined as follows:
\begin{equation}
    x = |\vec{v}_A - \vec{v}_B|_{2}-|\vec{v}_C - \vec{v}_D|_{2}
\end{equation}
Then we use the sentence analogy set created from \cite{zhu2020sentence}, which is specifically for this test. Here is a quadruple example:

\textit{A: A man is not singing.}

\textit{B: A man is singing.}

\textit{C: A girl is not playing the piano.}

\textit{D: A girl is playing the piano.}

We can see that the relation between A(C) and B(D) is negation; ideally, $x$ in this quadruple should be small. We list the performance of the sentence embedding method w/ and w/o Sen2Pro in Table~\ref{zzzz}.

\begin{table}[]\centering\small
\begin{tabular}{@{}cc|cc@{}}
\toprule
Embedding  & x  &Embedding  & x            \\ \midrule
BERT                & 21.3          &SBERT       & 15.6          \\
BERT+Ours           &\color[HTML]{3166FF}{14.7} &SBERT+Ours  & \color[HTML]{3166FF}{11.0} \\ \midrule
whitening      & 18.8          &SimCSE              & 12.0          \\
whitening+Ours & \color[HTML]{3166FF}{12.9} &SimCSE+Ours         & \color[HTML]{3166FF}{10.1} \\ \bottomrule
\end{tabular}
\caption{Results on the analogy task. Smaller $x$ means better. We can observe that our methods bring improvements to baselines.}
\label{zzzz}
\end{table}

\section{Conclusion and Future Work}
This paper investigates the probabilistic representation for sentences and proposes Sen2Pro, which portrays the representation uncertainty from model and data uncertainty. The effectiveness of Sen2Pro is theoretically explained and empirically verified through extensive experiments, which show the great potential of probabilistic sentence embedding. In the future, we will investigate several aspects of Sen2Pro, like how to pre-train language models from an uncertainty perspective since existing pre-training models are based on Sen2Vec. Also, we expect to design more natural schemes that utilize Sen2Pro instead of concatenating the mean and variance vectors. Such directions can further enhance Sen2Pro's performance and efficiency.

\section*{Limitation}
This major limitation of Sen2Pro lies in the computational cost due to the generation of several samples. Thus, improving the efficiency of Sen2Pro is a future direction. Besides, since we choose to concat the representation of different samples, there may be more natural ways to merge information from samples.

\newpage
\bibliography{acl_latex}
\bibliographystyle{acl_natbib}

\clearpage

\onecolumn
\appendix

\section{Proof of Theorem 1}\label{theorem1}
To accomplish further derivation on theorem 1. we first present a lemma from \cite{bickel2008regularized,bien2016convex}.
\begin{lemma}
Let $X_{i}$ be i.i.d. $N\left(0, \Sigma\right)$ and $\lambda_{\max }\left(\Sigma\right) \leq \epsilon$, where $\epsilon$ is a positive constant. Also, let $\sigma$ denote $\Sigma$, then we have
$$
P\left(\left|\sum_{i=1}^{n}\left(X_{i j} X_{i k}-\sigma_{j k}\right)\right| \geq n h\right) \leq C_{1} e^{-C_{2} n v^{2}} 
$$
for $|h| \leq \delta$, where $C_{1}, C_{2}$ and threshold $\delta$ depend on $\epsilon$ only, and $n$ is the sample number for $X_{i}$.
\end{lemma}

\begin{Proof}
We firstly denote the maximum element of a matrix $A$ as $|A|_{max}$, we have
\begin{gather}
\left\|B\left(\hat{\Sigma}\right)-B\left(\Sigma\right)\right\|_{2}\nonumber\\=O_{p}\left(\left\|B\left(\hat{\Sigma}\right)-B\left(\Sigma\right)\right\|_{\infty}\right)\nonumber\\=O_{p}\left(\left|B\left(\hat{\Sigma}\right)-B\left(\Sigma\right)\right|_{max}\right) \nonumber
\end{gather}
Naturally, based on Lemma 2, we can know that
$$
P\left(\left|B\left(\hat{\Sigma}\right)-B\left(\Sigma\right)\right|_{max} \geq t\right) \leq 3 k e^{-n t^{2} \gamma(\epsilon, \lambda)}
$$
for $|t| \leq \lambda(\epsilon) .$ By choosing $t=M(\log (k) / n)^{1 / 2}$, we conclude that 
$$
\left|B\left(\hat{\Sigma}\right)-\Sigma\right|_{max}=O_{p}\left(\left(n^{-1} \log k\right)^{1 / 2}\right)
$$
Putting them together, we have the result that completes the proof for Theorem 1.
\end{Proof}

\section{Theoretical Analysis}\label{theory}
This section presents theoretical explanations why Sen2Pro is better than Sen2Vec. 
Our framework is built on comparisons through distribution distances, which is inspired from representation learning based on KL divergence (relative entropy) \cite{chen2016infogan,li2019rate,gao2019rate,tschannen2019mutual}. 

\subsection{Problem Formulation}
We present the formulation of sentence representation: for a representation $X$ (with dimension $k$) of a sentence $s$, we assume that $X \sim P(\mu, \Sigma)$, where $P(\mu, \Sigma)$ is the ground-truth but unknown probability distribution. 

Recall that both Sen2Pro and Sen2Vec methods adopt the following representations:
\begin{itemize}
    \item Sen2Pro: sentence ${s}$ is represented as $(\hat{\mu},\hat{\Sigma})$, where $\hat{\mu} \in \mathcal{R}^{k \times 1}$ and $\hat{\Sigma}\in \mathcal{R}^{k \times 1}$ are estimated through i.i.d. samples $X_{1},X_{2}, \cdots, X_{N}$.   
    Therefore, Sen2Pro corresponds to a random variable $\hat{X} \sim P(\hat\mu,\hat\Sigma)$. 
    \item Sen2Vec: sentence ${s}$ is represented as $v \in \mathcal{R}^{k \times 1}$, where $v$ may be a sample $X_i$ from Sen2Pro.  
\end{itemize}
\paragraph{Probabilistic Viewpoint} 
Accordingly, from the viewpoint of probability, both Sen2Pro and Sen2Vec can be considered as the following random variables whose density functions are unknown but mean and covariance are known:
\begin{itemize}
    \item Sen2Pro: $X_{\text{P}}\sim P(\hat\mu, \hat\Sigma)$.
    \item Sen2Vec: $X_\text{V}\sim P(v, \epsilon I)$. 
\end{itemize}
where $\epsilon$ is a constant as a smoothing operator which makes a deterministic quantity $v$ like a random variable $X_\text{v}$. Note that since $\epsilon I$ is constant for any sentence ${s}$, $(v, \epsilon I)$ as sentence embedding is equivalent to $v$ as sentence embedding for downstream tasks. 

Then our goal is to show whether the $P(\hat\mu, \hat\Sigma)$ is closer to $P(\mu, \Sigma)$ than $P(v, \epsilon I)$, which is defined as follows:
\begin{equation}
    D(X_{\text{P}}, X) \textless D(X_{\text{V}}, X)
\end{equation}
Specifically, we use the KL divergence $D$ as the distance measurement.

\subsection{Superiority of Sen2Pro}\label{sec34}
In this part, we conduct derivations based on distribution distance to answer the above question. Unfortunately, since the density distributions of all three random variables $X$, $X_{\text{P}}$, $X_{\text{V}}$ are unknown, it is intractable to derive the equations on KL divergence. To this end, we make the following assumption about their density functions. 

\paragraph{Gaussian Distribution Assumption} 
Intuitively, an exponential family is suitable for the density function of Sen2Pro, such as Gaussian, Poisson, and Bernoulli. 
In this paper, we assume all three random variables $X$, $X_{\text{P}}$, $X_{\text{V}}$ are from Gaussian distribution, because non-Gaussian random variables possess smaller entropy (i.e., less uncertainty) than Gaussian random variables \cite{cover1999elements}.  
In this way, the ground-truth random variable $X$ contains the maximal uncertainty.

Based on the above assumption, we have: $X\sim \mathcal{N}(\mu, \Sigma)$, $X_{\text{P}} \sim \mathcal{N}(\hat \mu, \hat \Sigma)$, and $X_{\text{V}} \sim \mathcal{N}(\hat v, \epsilon I)$. 
Then, we provide the following theoretical guarantee to show the superiority of Sen2Pro, which indicates that $\mathcal{N}(\hat\mu, \hat\Sigma)$ is closer to $\mathcal{N}(\mu, \Sigma)$ than $\mathcal{N}(v, \epsilon I)$: 
\begin{theorem}\label{th3}
If $\epsilon \textless \frac{\sqrt[k]{|\hat{\Sigma}|}}{e}$ with $e$ as the natural number, then the following inequality holds:
\begin{equation}
    D(X_{\text{P}}, X) \textless D(X_{\text{V}}, X)
\end{equation}
\end{theorem}

\begin{figure}
    \centering
    \includegraphics[scale=0.23]{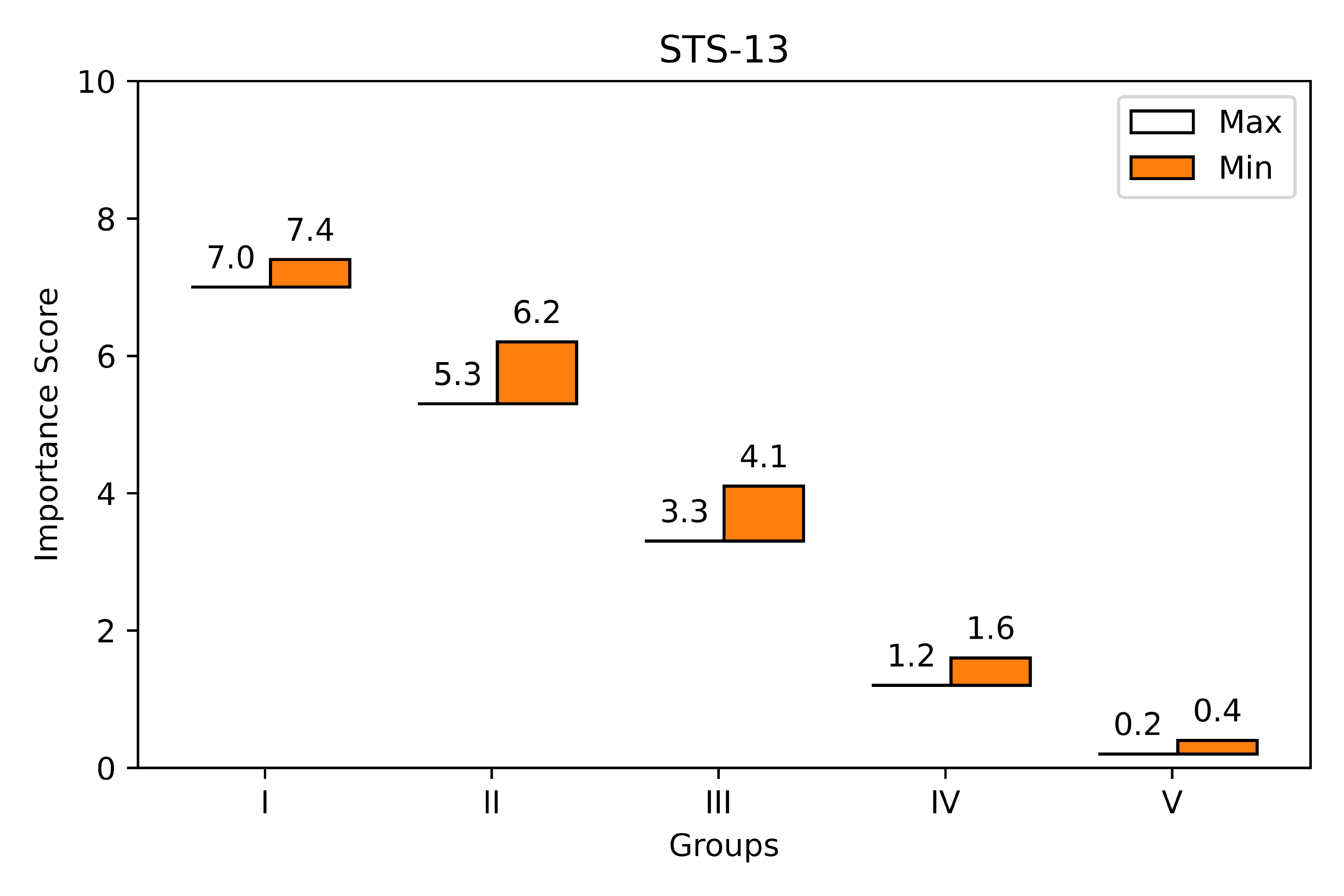}
    \includegraphics[scale=0.23]{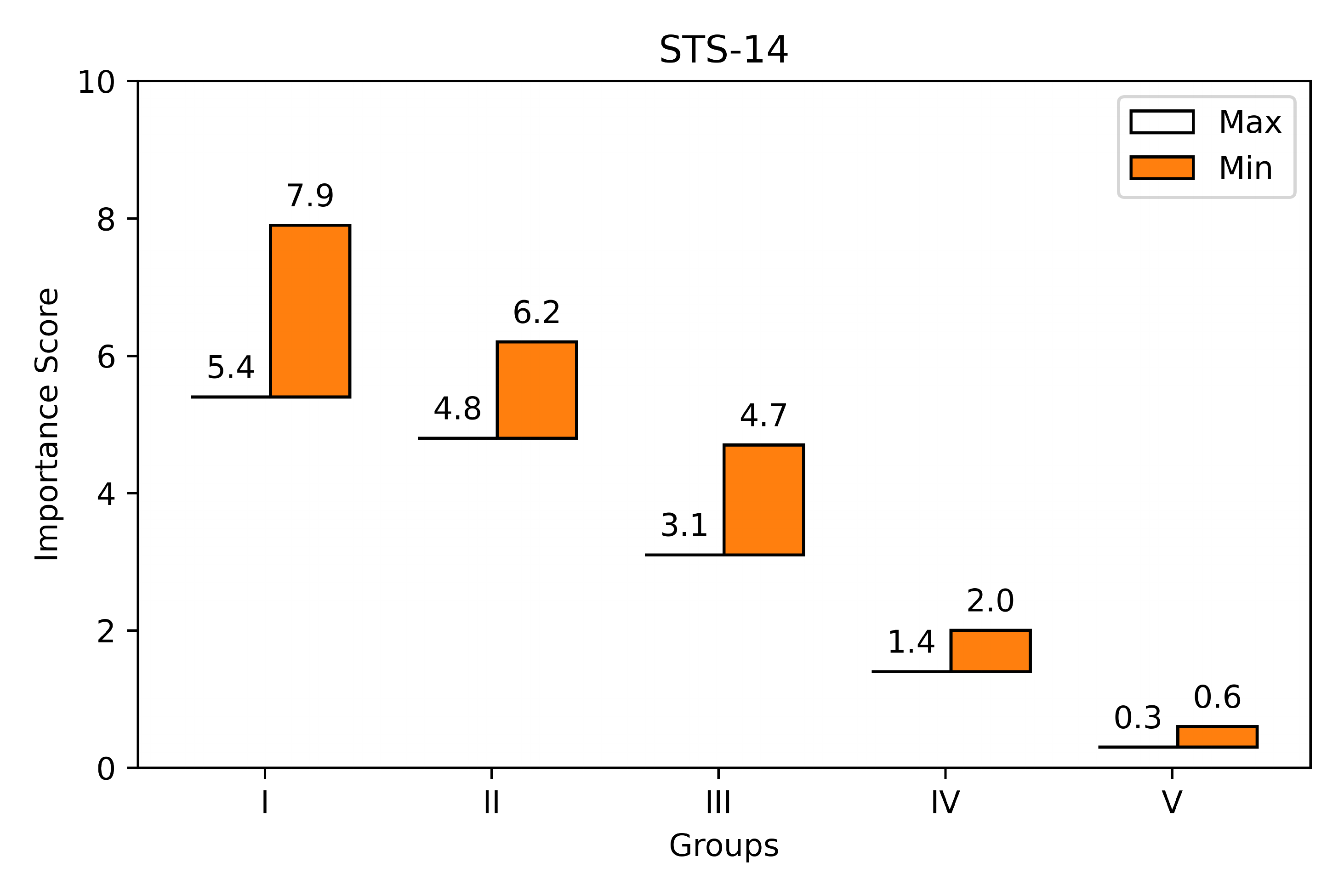}
    \\
    \includegraphics[scale=0.23]{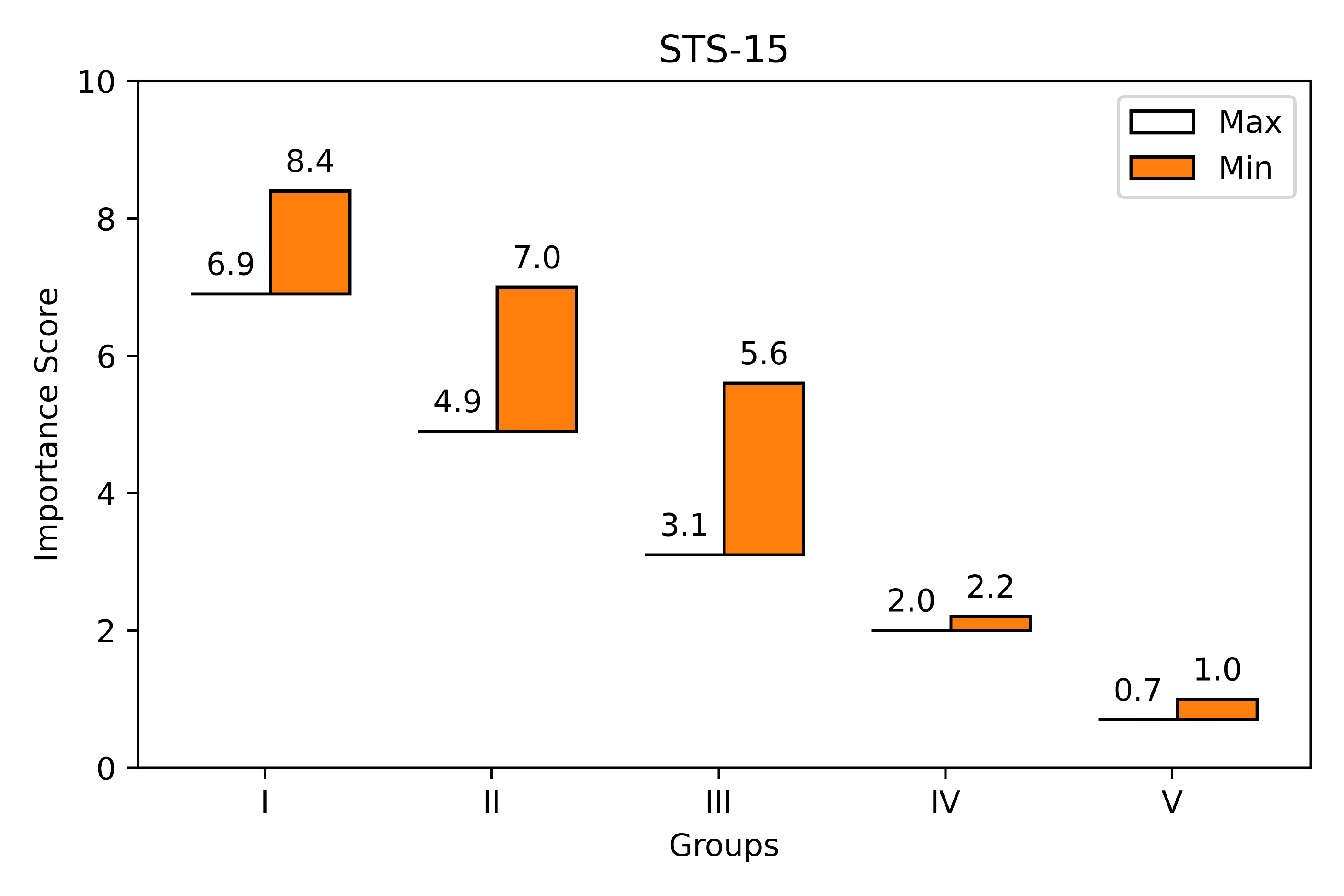}
    \includegraphics[scale=0.23]{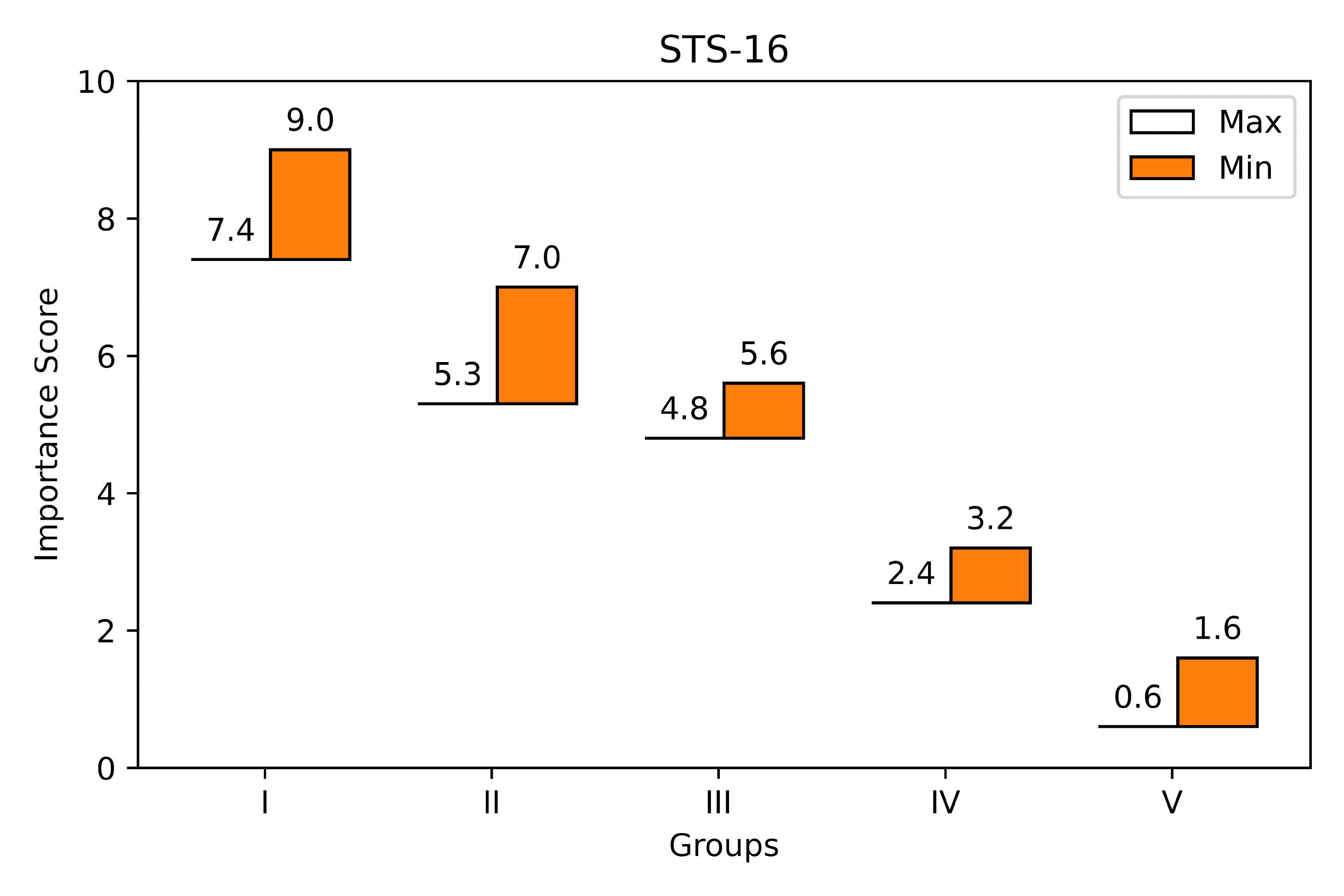}
    \\
    \includegraphics[scale=0.23]{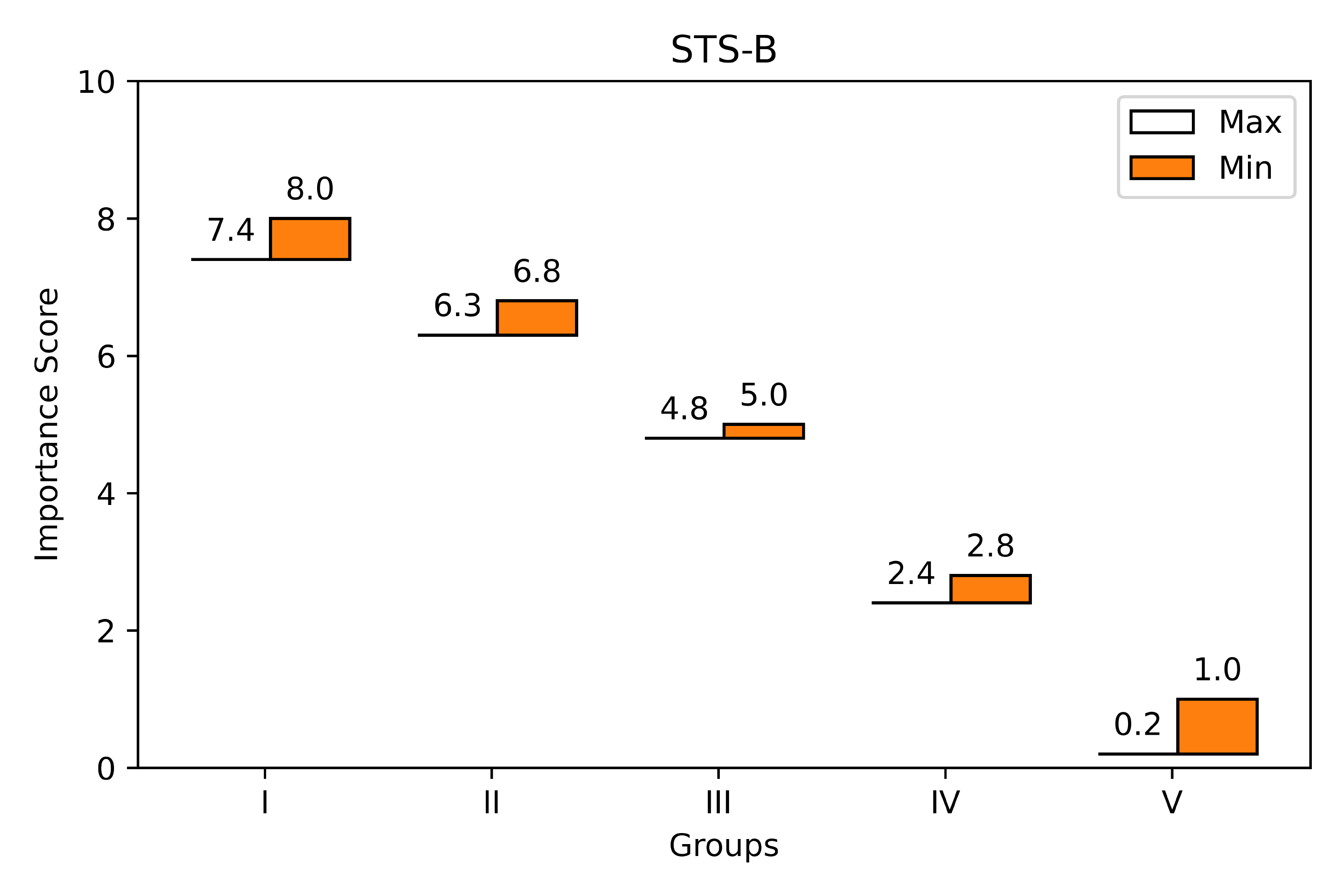}
    \includegraphics[scale=0.23]{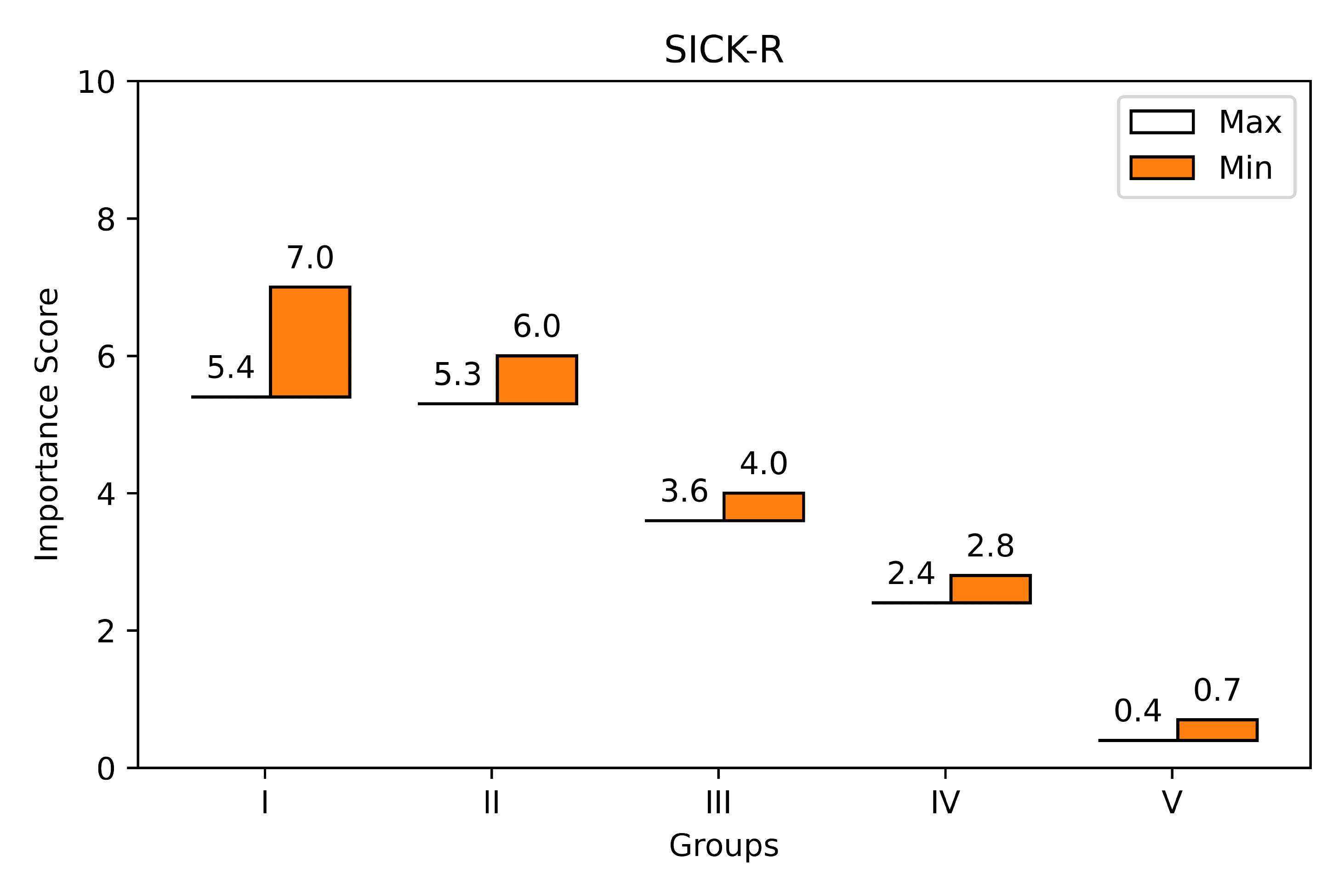}
    \caption{The importance score of five groups of features on STS benchmarks.}
    \label{c}
\end{figure}

\section{Proof of Theorem 2}
Before we start to prove the theorem, we provide some definitions that will be useful for further derivation:
\begin{definition}
For a random variable $X$ with mean $\mu$, the moment generating function is
\begin{gather}
    M_{X-\mu}(\lambda)=\mathbb{E} \exp (\lambda(X-\mathbb{E} X))
\end{gather}

the cumulant-generating function is
\begin{gather}
    \Gamma_{X-\mu}(\lambda)=\log M_{X-\mu}(\lambda)
\end{gather}
\end{definition}

We denote $D(X_{\text{P}}, X)$ and $ D(X_{\text{V}}, X)$ as $D_{a}$ and $D_{b}$, and rewrite $D(X_{\text{P}}, X)$ and $ D(X_{\text{V}}, X)$ as follows:
\begin{equation}
    D(\mathcal{N}(\mu, \Sigma) \| \mathcal{N}(\hat \mu, \hat \Sigma)) \\=  \frac{1}{2}[\log ( \frac{|\Sigma|}{|\epsilon I|})+\operatorname{tr}\left(\epsilon I \Sigma^{-1}\right)-k]
\end{equation}
\begin{equation}
D(\mathcal{N}(\mu, \Sigma) \| \mathcal{N}(\hat v, \epsilon I)) \\= \frac{1}{2}[\log ( \frac{|\Sigma|}{|\hat{\Sigma}|})+\operatorname{tr}\left(\Sigma \hat{\Sigma}^{-1}\right)-k\\+(\mu-\hat{\mu})^{\top} \hat{\Sigma}^{-1}(\mu-\hat{\mu})]
\end{equation}

\begin{Proof}
As defined in the main paper, we have $D_{a}=D(\mathcal{N}(\mu, \Sigma) \| \mathcal{N}(\hat \mu, \hat \Sigma)) =  \frac{1}{2}[\log ( \frac{|\Sigma|}{|\epsilon I|})+\operatorname{tr}\left(\epsilon I \Sigma^{-1}\right)-p]
$ and $D_{b}=D(\mathcal{N}(\mu, \Sigma) \| \mathcal{N}(\hat v, \epsilon I)) = \frac{1}{2}[\log ( \frac{|\Sigma|}{|\hat{\Sigma}|})+\operatorname{tr}\left(\Sigma \hat{\Sigma}^{-1}\right)-p+(\mu-\hat{\mu})^{\top} \hat{\Sigma}^{-1}(\mu-\hat{\mu})]
$. As illustrated in the methodology of Sen2Pro, we have the following derivation:
\begin{equation}
|\Sigma^{-1}|=\prod_{i=1}^{n} \frac{1}{\Sigma_{ii}}
\end{equation}
\begin{equation}
|\Sigma|=\prod_{i=1}^{n} \Sigma_{ii}
\end{equation}

Then we focus on the term:

\begin{equation}
\begin{aligned}
    D_{a}&=D(\mathcal{N}(\mu, \Sigma) \| \mathcal{N}(\hat \mu, \hat \Sigma)) \\&=  \frac{1}{2}[\log ( \frac{|\Sigma|}{|\epsilon I|})+\operatorname{tr}\left(\epsilon I \Sigma^{-1}\right)-p]\\
&\approx\frac{1}{2}[\log ( \frac{|\Sigma|}{|\epsilon I|})-p]\\
\end{aligned}
\end{equation}
since $\epsilon$ is an extremely small number. Besides, we also make derivation on $D_{b}$:
\begin{equation}
\begin{aligned}
    D_{a}&=D(\mathcal{N}(\mu, \Sigma) \| \mathcal{N}(\hat \mu, \hat \Sigma)) \\&=  \frac{1}{2}[\log ( \frac{|\Sigma|}{|\hat{\Sigma}|})+\operatorname{tr}\left(\Sigma \hat{\Sigma}^{-1}\right)-p\\&+(\mu-\hat{\mu})^{\top} \hat{\Sigma}^{-1}(\mu-\hat{\mu})]\\
&=\frac{1}{2}[\log ( \frac{|\Sigma|}{|\hat{\Sigma}|})+(\mu-\hat{\mu})^{\top} \hat{\Sigma}^{-1}(\mu-\hat{\mu})]\\
\end{aligned}
\end{equation}
To better analyze the $(\mu-\hat{\mu})^{\top} \hat{\Sigma}^{-1}(\mu-\hat{\mu})$ term of $D_{a}$, we present the following lemma:
\begin{lemma}
For $X \sim N\left(\mu, \sigma^{2}\right)$, where $X$ is a single Gaussian variable,
$$
\Gamma_{X-\mu}(\lambda)=\frac{\lambda^{2} \sigma^{2}}{2}, \quad \quad \Gamma_{X-\mu}^{*}(\epsilon)=\frac{\epsilon^{2}}{2 \sigma^{2}} .
$$
For $X_{1}, \ldots, X_{n} \sim N\left(\mu, \sigma^{2}\right)$, it's easy to check that the bound is tight:
$$
\lim _{n \rightarrow \infty} \frac{1}{n} \ln P\left(\bar{X}_{n}-\mu \geq \epsilon\right)=-\frac{\epsilon^{2}}{2 \sigma^{2}} .
$$
\end{lemma}

Therefore, we can represent $|\mu-\hat{\mu}|_{2}$ as $O(\epsilon)$ and we have the following:
\begin{equation}
\begin{aligned}
    &(\mu-\hat{\mu})^{\top} \hat{\Sigma}^{-1}(\mu-\hat{\mu})\\&=D(\mathcal{N}(\mu, \Sigma) \| \mathcal{N}(\hat \mu, \hat \Sigma)) \\&=  O(\epsilon^2)\\
\end{aligned}
\end{equation}
Naturally, we drop all $O(\epsilon^2)$ term and formulate $D_{a}$ as follows:
\begin{gather}
    D_{a}=\frac{1}{2}[\log ( \frac{|\Sigma|}{|\hat{\Sigma}|})+\operatorname{tr}\left(\Sigma \hat{\Sigma}^{-1}\right)-p]\\=\frac{1}{2}[\log ( \frac{|\Sigma|}{|\hat{\Sigma}|})]
\end{gather}
Then we insert the condition to the equality $D_{a} < D_{b}$, and we can complete the proof.
\end{Proof}

\section{Comparisons between individual estimation and unified estimation}\label{compare}
This section conducts comparisons between individual estimation and unified estimation. 
\begin{itemize}
    \item Concretely, individual estimation refers to estimate $\mu^m, \hat \Sigma^m$ and $\mu^d,\hat \Sigma^d$ based on $\mathcal{S}^m$ and $\mathcal{S}^d$, respectively. Then we use $\frac{\mu^m + \mu^d}{2}$ and $\frac{\hat \Sigma^m + \hat \Sigma^d}{2}$ for sentence representation.
    \item Correspondingly, unified estimation refers to estimate $\mu^u , \hat \Sigma^u$ based on $\mathcal{S}^m \cap \mathcal{S}^d$. Then use $\mu^u , \hat \Sigma^u$ for sentence representation.
\end{itemize}

The experimental settings follow the Sec~\ref{Experiment}, and the results are shown in Table~\ref{compare1}. We can observe that unified estimation under-performs individual estimation. Thus, we choose to estimate uncertainty in an individual way.

\begin{table*}[!h]\small
\centering
\begin{tabular}{@{}c|ccccccc|c@{}}
\toprule
Model            & STS-12 & STS-13 & STS-14 & STS-15 & STS-16 & STS-B & SICK-R & Average          \\ \midrule
BERT-base        & 57.84  & 61.95  & 62.48  & 70.95  & 69.81  & 59.04 & 63.75  & 63.69        \\
BERT-base-$G_{m+d}$     & \color[HTML]{3166FF}59.40  & \color[HTML]{3166FF}63.03  & \color[HTML]{3166FF}64.18  & \color[HTML]{3166FF}71.97  & \color[HTML]{3166FF}70.73  & \color[HTML]{3166FF}62.59 & \color[HTML]{3166FF}64.69  & \color[HTML]{3166FF}65.23 (+1.54) \\
BERT-base-$G_{u}$      & 58.86  & 62.66  & 63.77  & 71.48  & 70.56  & 61.89 & 64.28  & 64.79 (+1.02) \\ \bottomrule
\end{tabular}
\caption{BERT-base-$G_{m+d}$ and BERT-base-$G_{u}$ represent Sen2Pro with individual and unified estimation, respectively.}
\label{compare1}
\end{table*}

\section{Relation between model uncertainty and feature importance}\label{importance}
This section serves as a complementary material for demonstrating the relation between the model uncertainty and feature importance, and the results on the STS task are illustrated in the Figure~\ref{a} \ref{b} \ref{c}.

\section{Relation between model uncertainty and performance improvements}\label{performance}

\begin{figure*}
    \centering
    \includegraphics[scale=0.43]{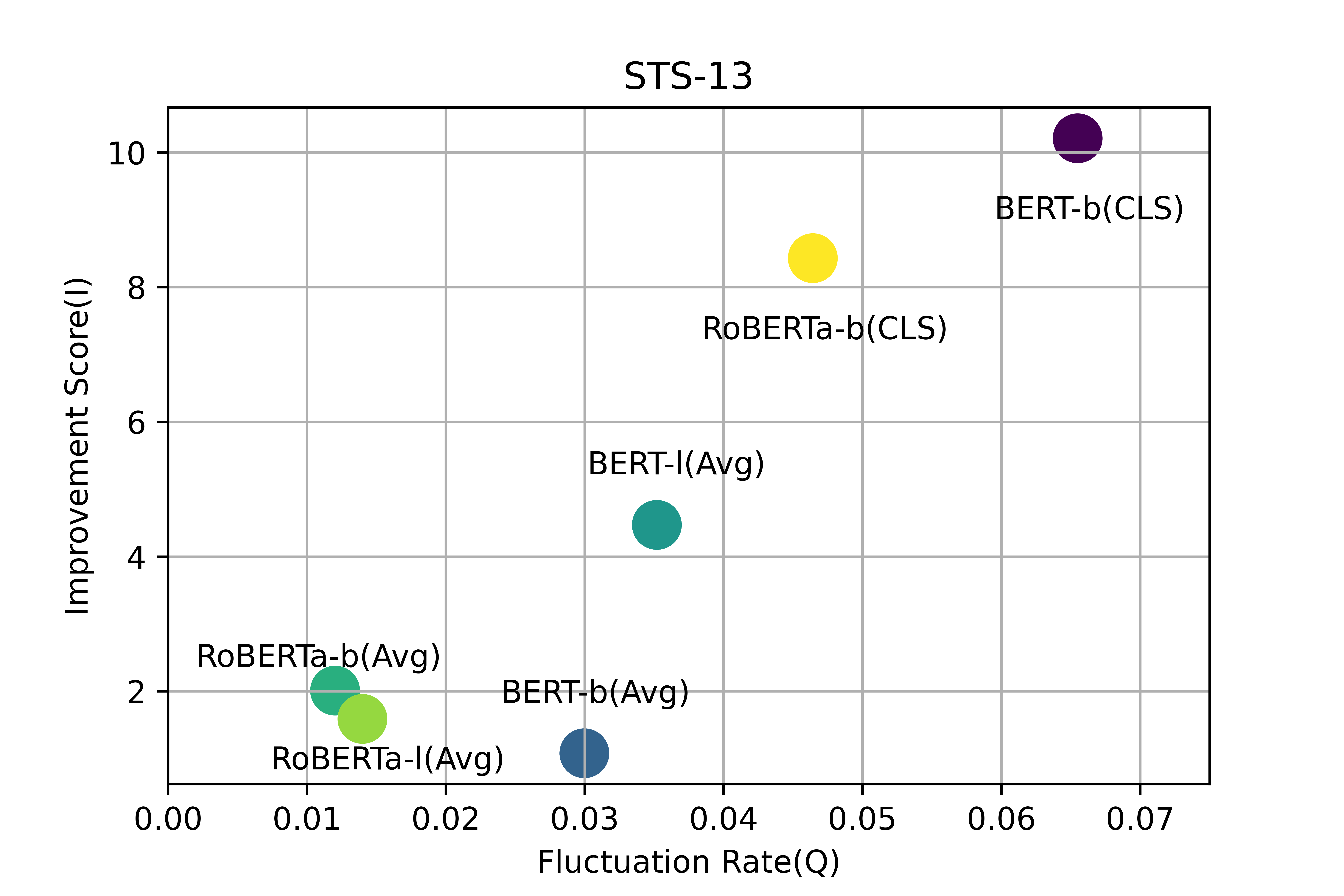}
    \includegraphics[scale=0.43]{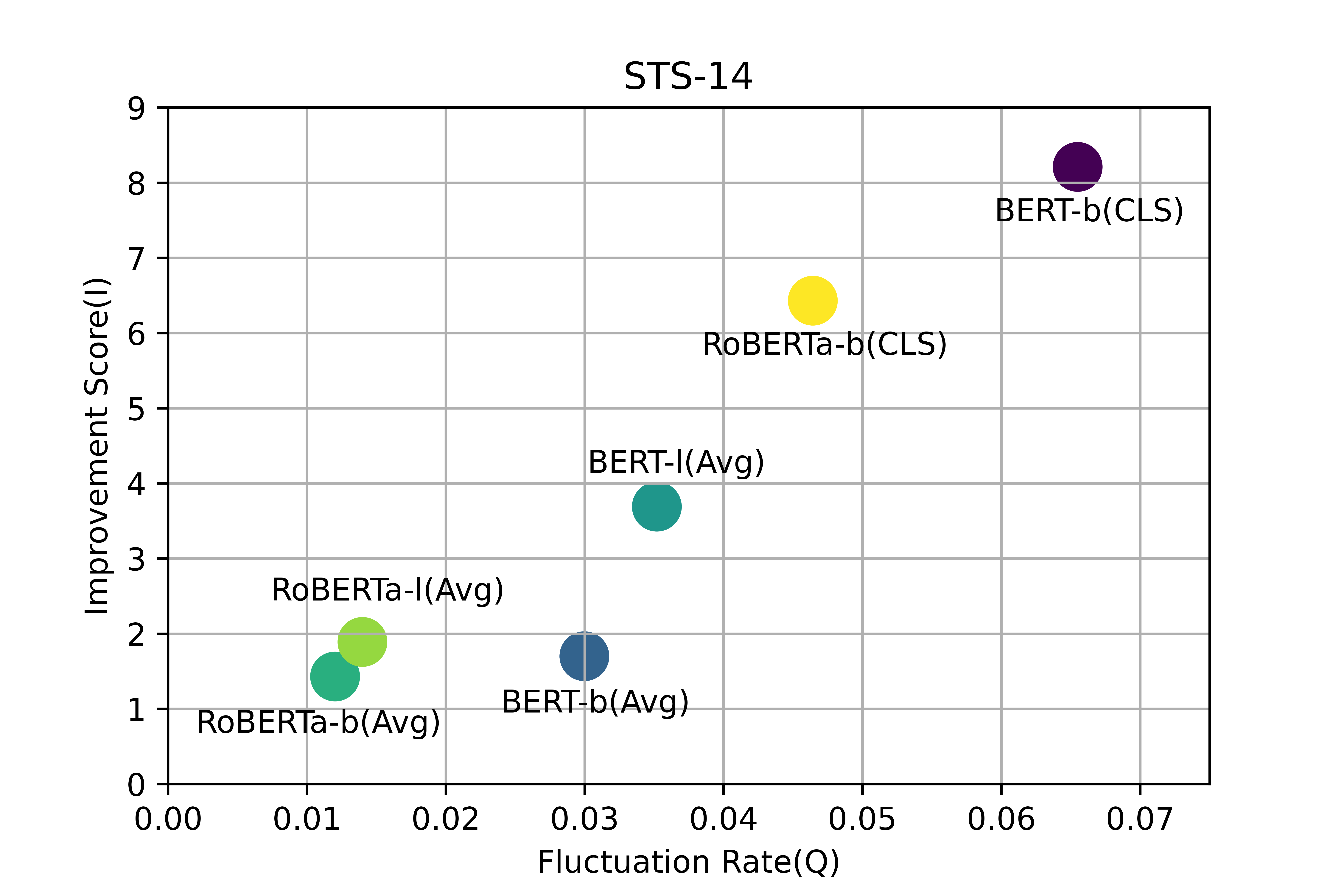}
    \\
    \includegraphics[scale=0.43]{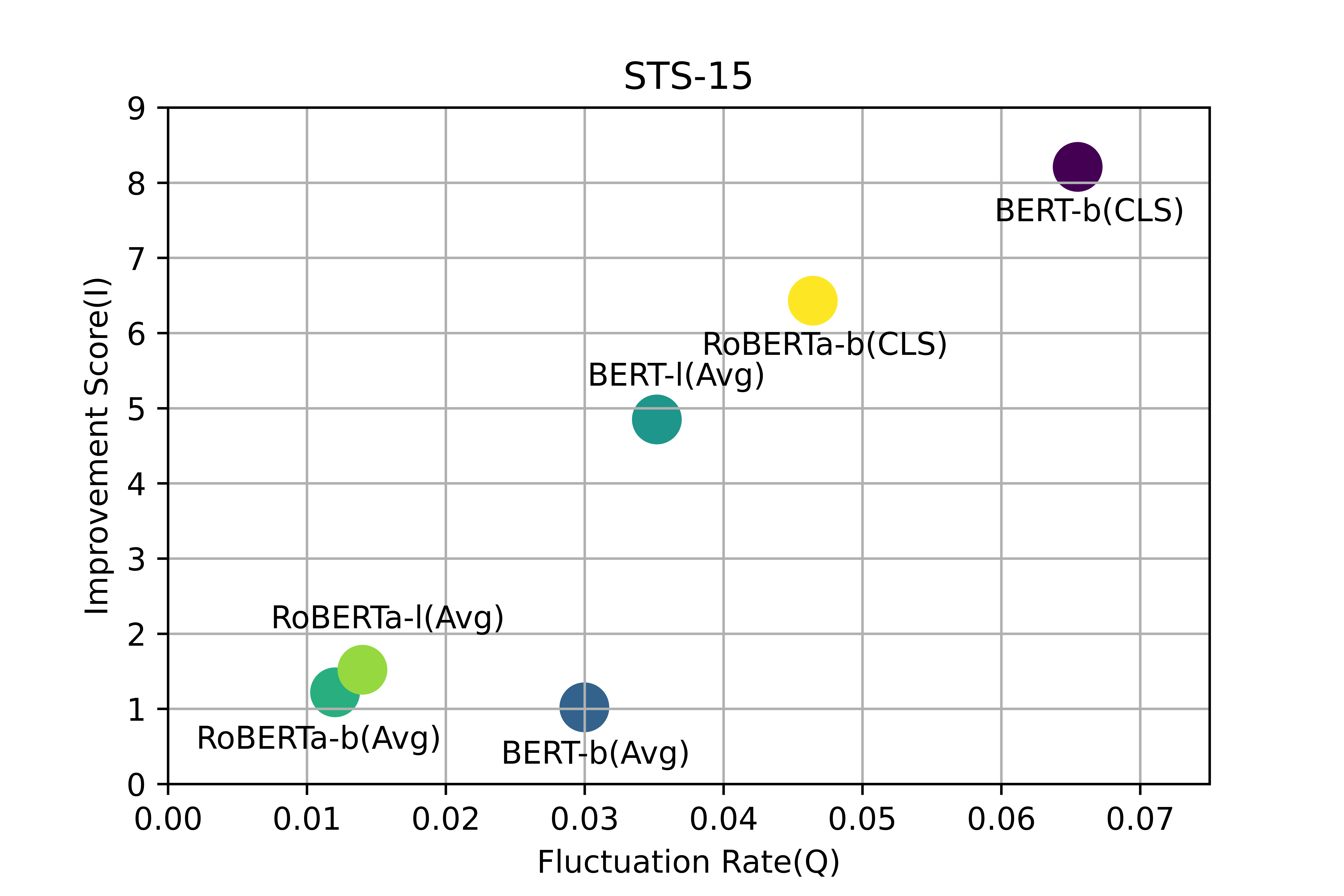}
    \includegraphics[scale=0.43]{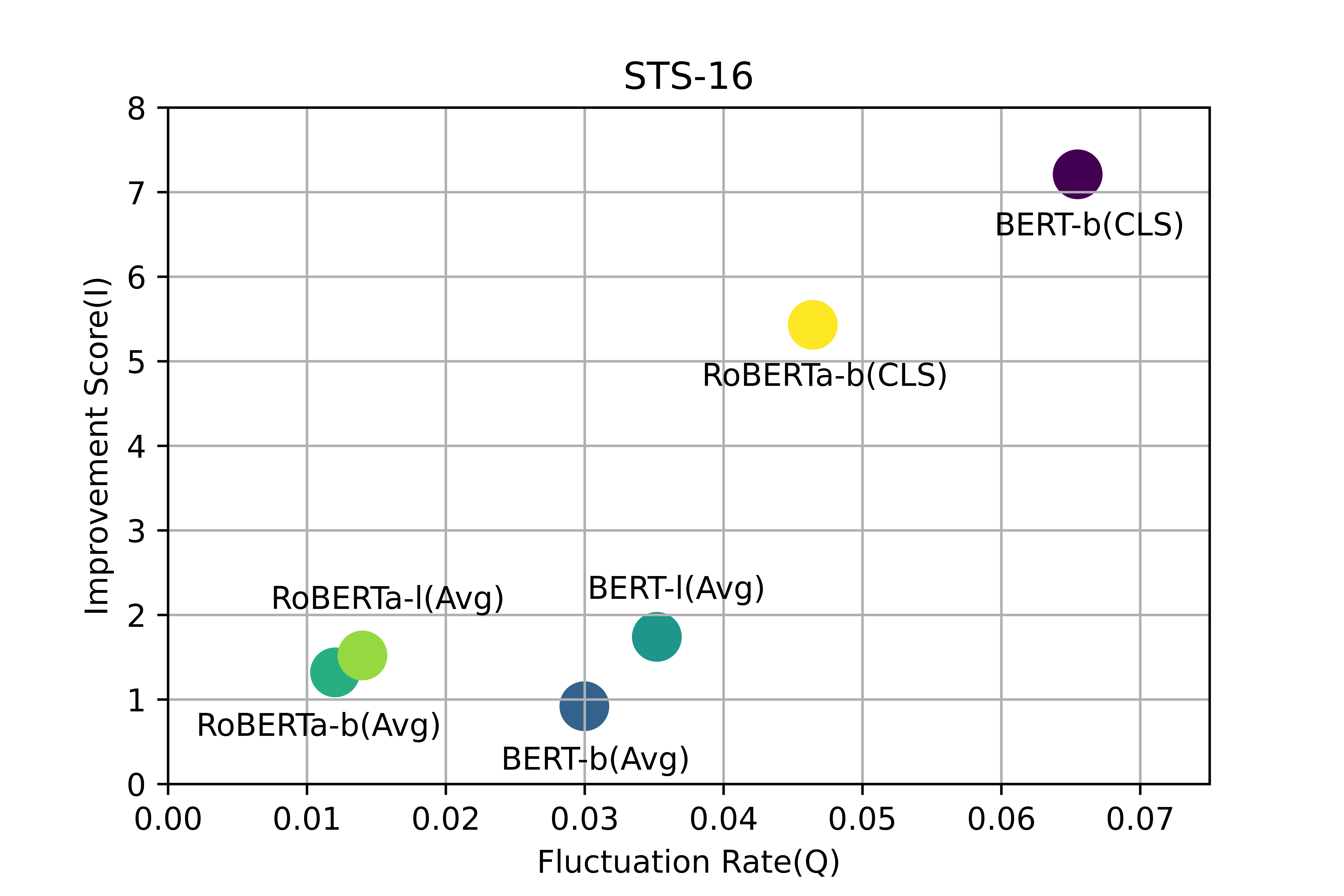}
    \\
    \includegraphics[scale=0.43]{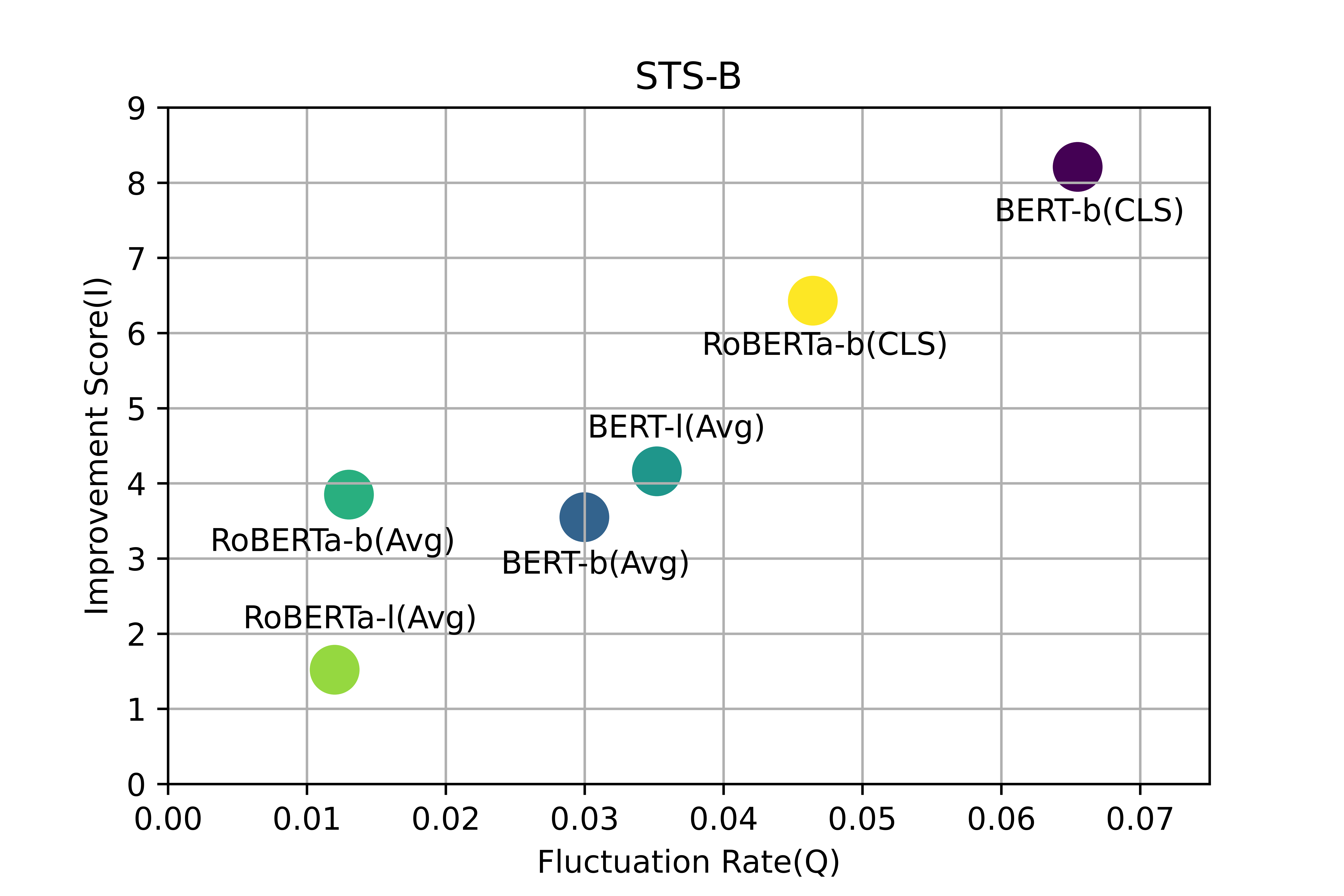}
    \includegraphics[scale=0.43]{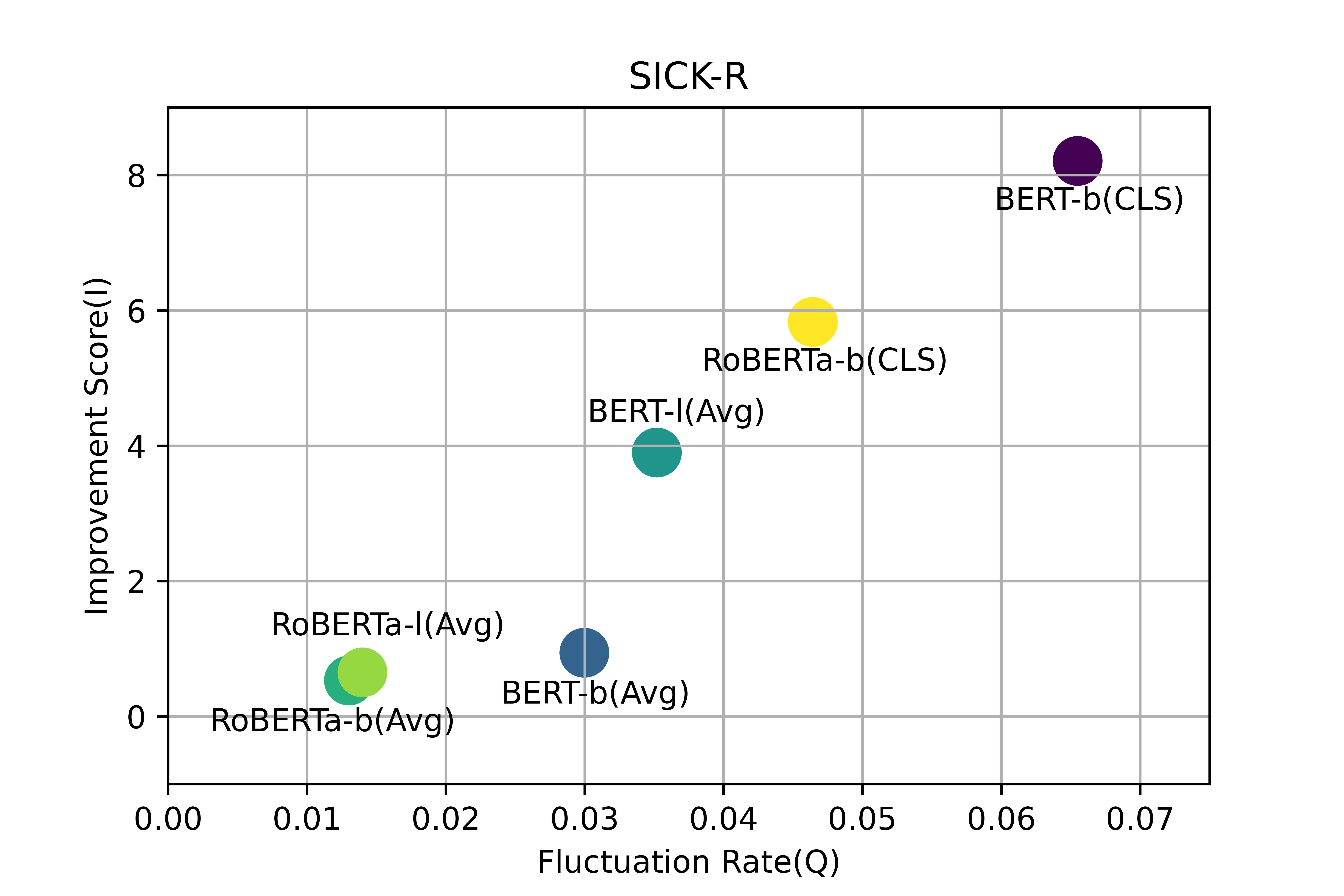}
    \caption{The relation between fluctuation rate ($Q$) and improvement score ($I$) on STS benchmarks.}
    \label{fig:6}
\end{figure*}

This section serves as a complementary material for demonstrating the relation between the model uncertainty and performance improvements, and the results on the STS task are illustrated in the Figures \ref{fig:6}.

\section{Influence of the Sampling Number}\label{ablation:model}
\begin{figure*}
    \centering
    \includegraphics[scale=0.23]{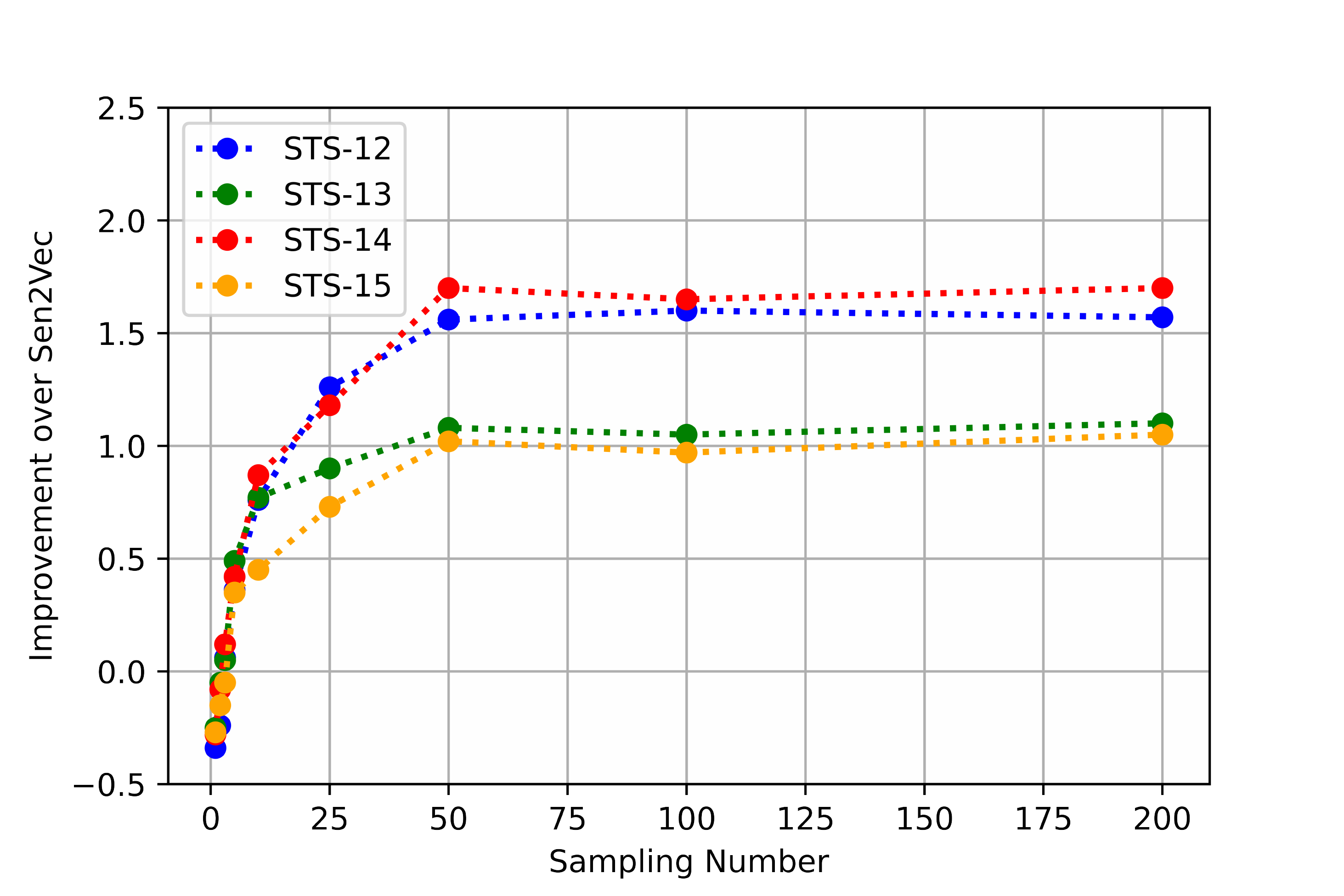}
    \includegraphics[scale=0.23]{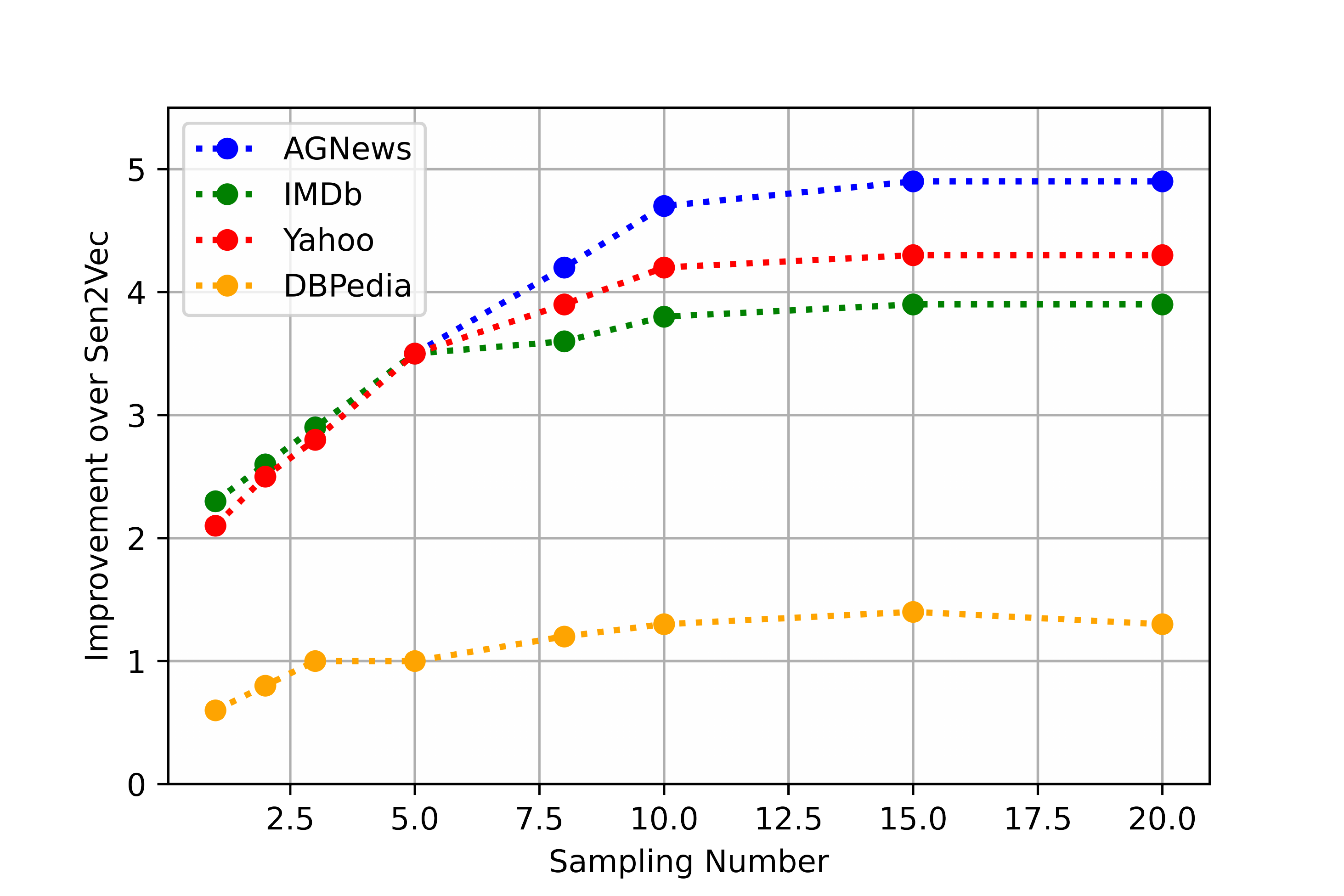}
    \caption{Results as the sampling number $n_{1}$ changes with `BERT-base-G'. The text classification task uses the `10-shot' setting.}
    \label{fig:3}
\end{figure*}

\begin{figure*}
    \centering
    \includegraphics[scale=0.46]{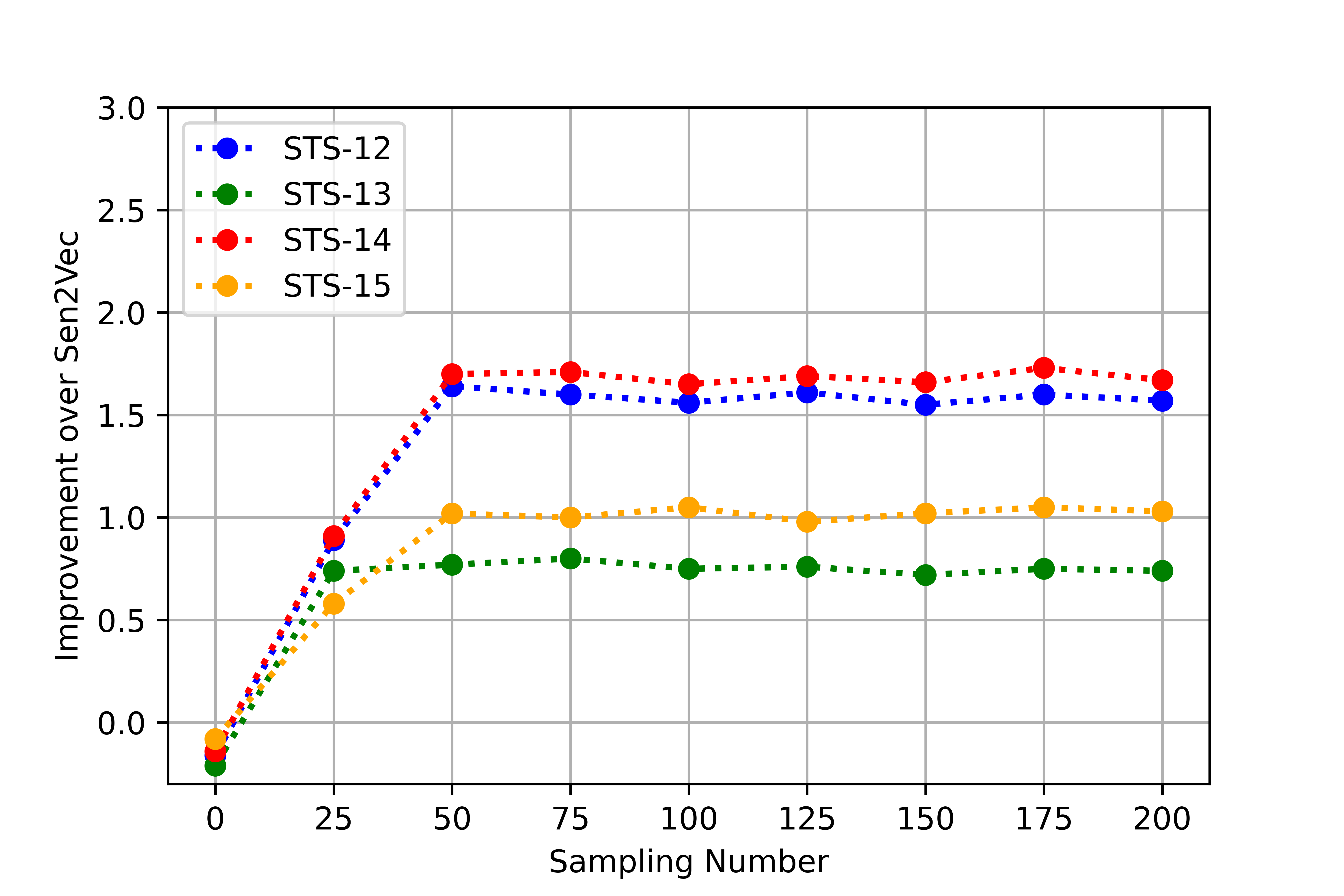}
    \includegraphics[scale=0.46]{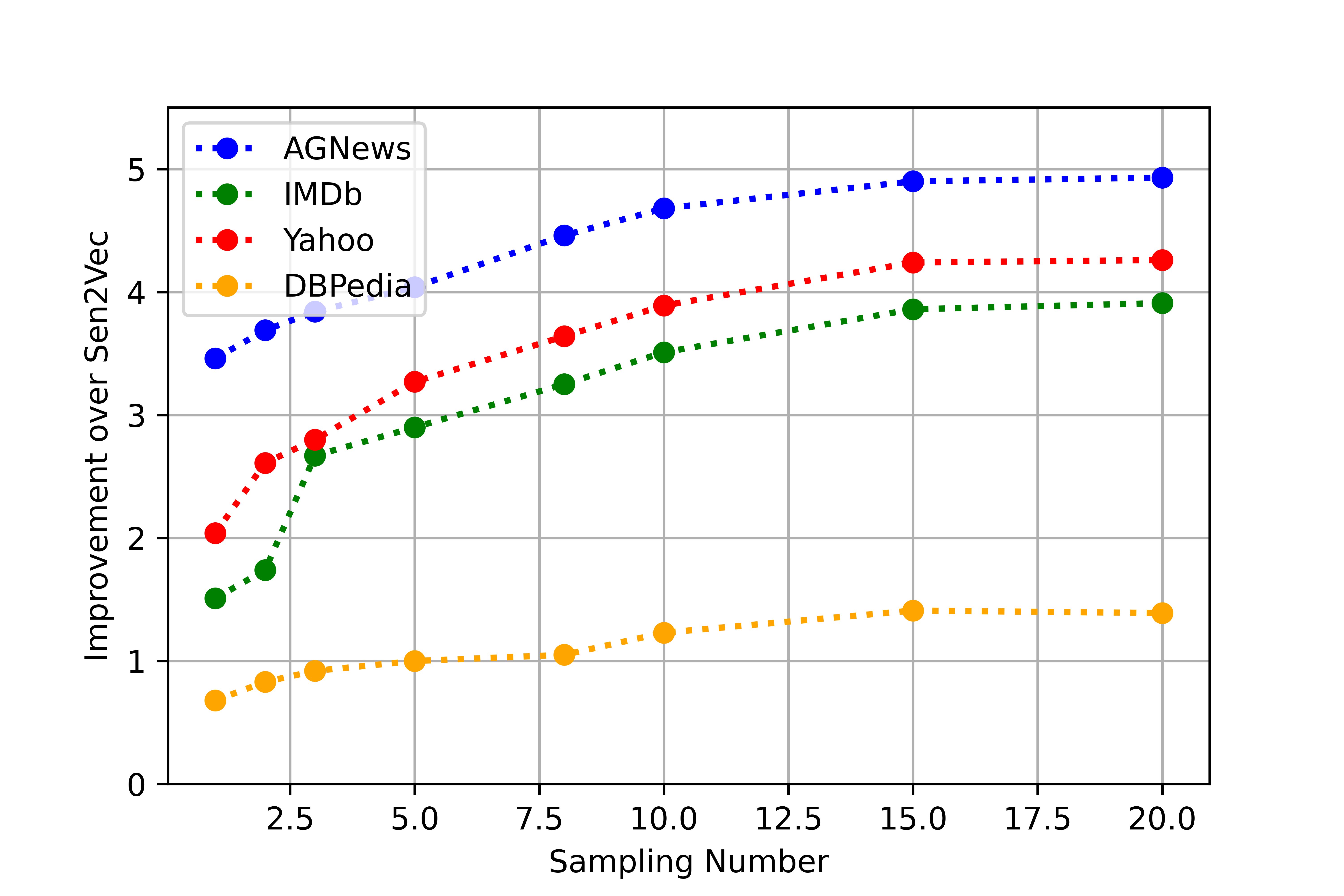}
    \caption{Results as sampling number $n_{2}$ changes with `BERT-base-G'. The text classification task uses the `10-shot' setting.}
    \label{ssssss}
\end{figure*}

This section examines the effect of the sampling number $n_{1}$ in the model uncertainty and the sampling number $n_{2}$. To examine the effect of $n_{1}$, an experiment is run, where $n_{1}$ varies and $n_{2}$ is set $30$.  The results of the STS and text classification task are illustrated in Figure~\ref{fig:3}. From Figure~\ref{fig:3}, we select $n_{1}=50$ and $n_{1}=15$ for the STS and text classification task, respectively, since smaller $n_{1}$ result in imprecise sentence representation and larger $n_{1}$ brings significant computational burden with little improvements. To examine the effect of the sampling number $n_{2}$, we change $n_{2}$ and keep the optimal $n_{1}$, and show the results in Figure~\ref{ssssss}. Overall, when $n_{2}$ is greater than 30, Sen2Pro achieves a stable performance.  

\section{Results of Segment-level Evaluation of Neural Machine Translation}\label{segment}
In the main paper, we make the evaluations on the system-level NMT. Here, we present the results of segment-level Evaluation of Neural Machine Translation. The experimental settings are kept the same as the ones in our main paper. The results are listed in Table~\ref{tabnmt}. As the results demonstrate, Sen2Pro outperforms Sen2Vec and achieves comparable performances to BERTScore.

\begin{table*}[!h]\large
\centering
\begin{tabular}{@{}c|ccccccc|c@{}}
\toprule
Metric    & cs-en                        & de-en                        & fi-en                        & lv-en                        & ru-en                        & tr-en                        & zh-en                        & Avg                          \\ \midrule
BLEU      & 23.3                        & 41.5                        & 28.5                        & 15.4                        & 22.8                        & 14.5                        & 17.8                        & 21.2                        \\
ITER      & 19.8                        & 39.6                        & 23.5                        & 12.8                        & 13.9                        & 2.9                        & 14.4                        & 18.1                        \\
RUSE     & 34.7                        & 49.8                        & 36.8                        & 27.3                        & 31.1 & 25.9                        & 21.8                        & 32.5                        \\
Sen2Vec   & 33.8                        & 48.6                        & 35.9                        & 26.0                        & 29.0                        & 22.4                        & 21.0                        & 31.0                        \\
BERTScore  & 36.5 & 51.4                        & 37.2 & 26.8 & 31.7                        & 27.2&  22.9                        & 33.4                       \\
Sen2Pro   & 38.7                        & 54.1 & 38.9                        & 28.3                        & 34.5                        & 28.0                        & 24.8 & 35.3 \\ \bottomrule
\end{tabular}
\caption{Pearson correlations with segment-level machine translation evaluation on WMT17.}
\label{tabnmt}
\end{table*}

\end{document}